\documentclass[journal]{IEEEtran}

\usepackage{amsmath}    % 用于高级数学公式
\usepackage{amssymb}    % 用于数学符号
\usepackage{graphicx}   % 用于插入图片
\usepackage{longtable}
\usepackage{booktabs}
\usepackage{threeparttable} % 提供表注环境
\usepackage{caption}
\usepackage{subcaption} 
\usepackage{hyperref}
\usepackage{csquotes} % for \enquote
\hypersetup{
    colorlinks=true,
    linkcolor=red,
    filecolor=magenta,      
    urlcolor=cyan,
    citecolor=green,
}
\usepackage[linesnumbered,ruled]{algorithm2e}
\usepackage{float}  % 用于强制表格和算法位置
\usepackage{placeins}  % 提供 \FloatBarrier 命令
\usepackage[switch]{lineno}
\ifCLASSINFOpdf
\else
\fi
\begin{document}
% \title{Robust Tiny Object Detection in Remote Sensing Images via MoE and Density-Based Guidance}
% \title{Bridging the Scale Gap: Balanced Tiny-to-Large Object Detection in Remote Sensing Imagery}

\title{Bridging the Scale Gap: Balanced Tiny and General Object Detection in Remote Sensing Imagery}



\author{Zhicheng Zhao, Yin Huang, Lingma Sun, Chenglong Li, Jin Tang
% <-this % stops a space

\thanks{This work was supported in part by the National Natural Science Foundation of China(No. 62306005, 62006002, and 62076003), and in part by the Natural Science Foundation of Anhui Higher Education Institution (No. 2022AH040014). (Corresponding author: Lingma Sun).}% <-this % stops a space
\thanks{Zhicheng Zhao, Yin Huang, and Chenglong Li are with Key Laboratory of Intelligent Computing \& Signal Processing (Anhui University), Ministry of Education, Anhui Provincial Key Laboratory of Multimodal Cognitive Computation, School of Artificial Intelligence, Anhui University, Hefei 230601, China. Zhicheng Zhao is also with the 38th Research Institute, China Electronics Technology Group Corporation, Hefei 230088, China. (Email: zhaozhicheng@ahu.edu.cn, wa23301134@stu.ahu.edu.cn, lcl1314@foxmail.com).}
\thanks{Lingma Sun is with Collaborative Innovation Laboratory for Computer Vision and Pattern Recognition, School of Artificial Intelligence and Big Data, Hefei University, Hefei 230601, China. (Email: sunlm@hfuu.edu.cn).}
\thanks{Jin Tang is with Anhui Provincial Key Laboratory of Multimodal Cognitive Computation, School of Computer Science and Technology, Anhui University, Hefei 230601, China. (Email: tangjin@ahu.edu.cn).}}% <-this % stops a space

\markboth{Journal of \LaTeX\ Class Files,~Vol.~13, No.~9, September~2014}%
{Shell \MakeLowercase{\textit{et al.}}: MoE Tiny Detr}

\maketitle 

\begin{abstract}

Tiny object detection in remote sensing imagery has attracted significant research interest in recent years. Despite recent progress, achieving balanced detection performance across diverse object scales remains a formidable challenge, particularly in scenarios where dense tiny objects and large objects coexist. Although large foundation models have revolutionized general vision tasks, their application to tiny object detection remains unexplored due to the extreme scale variation and density distribution inherent to remote sensing imagery. To bridge this scale gap, we propose ScaleBridge-Det, to the best of our knowledge, the first large detection framework designed for tiny objects, which could achieve balanced performance across diverse scales through scale-adaptive expert routing and density-guided query allocation. Specifically, we introduce a Routing-Enhanced Mixture Attention (REM) module that dynamically selects and fuses scale-specific expert features via adaptive routing to address the tendency of standard MoE models to favor dominant scales. REM generates complementary and discriminative multi-scale representations suitable for both tiny and large objects. Furthermore, we present a Density-Guided Dynamic Query (DGQ) module that predicts object density to adaptively adjust query positions and numbers, enabling efficient resource allocation for objects of varying scales. The proposed framework allows ScaleBridge-Det to simultaneously optimize performance for both dense tiny and general objects without trade-offs. Extensive experiments on benchmark and cross-domain datasets demonstrate that ScaleBridge-Det achieves state-of-the-art performance on AI-TOD-V2 and DTOD, while exhibiting superior cross-domain robustness on VisDrone.
\end{abstract}


% \fi
\IEEEpeerreviewmaketitle

\section{INTRODUCTION}

\IEEEPARstart{R}{emote} sensing object detection is a fundamental task in computer vision and serves as a critical component in various civilian applications, including environmental monitoring, urban construction, disaster response, and surveillance \cite{wu2020detection,li2020uav,zhang2020disaster,chen2020surveillance}. The rapid development of unmanned aerial vehicle (UAV) and satellite imaging technologies has created a pressing need to automatically detect objects across diverse scales. Remote sensing images naturally exhibit large scale variations, spanning from tiny objects such as small vehicles or pedestrians, which occupy only a few pixels, to large structures like buildings and bridges that cover hundreds of pixels \cite{li2023smallobj,liu2023domain} (depicted in Fig. 1). Furthermore, multi-scale targets commonly coexist within the same scene, posing significant challenges to current detection systems.

\begin{figure}[t]
\centering
\begin{subfigure}[b]{0.48\columnwidth}
    \centering
    \includegraphics[width=\textwidth]{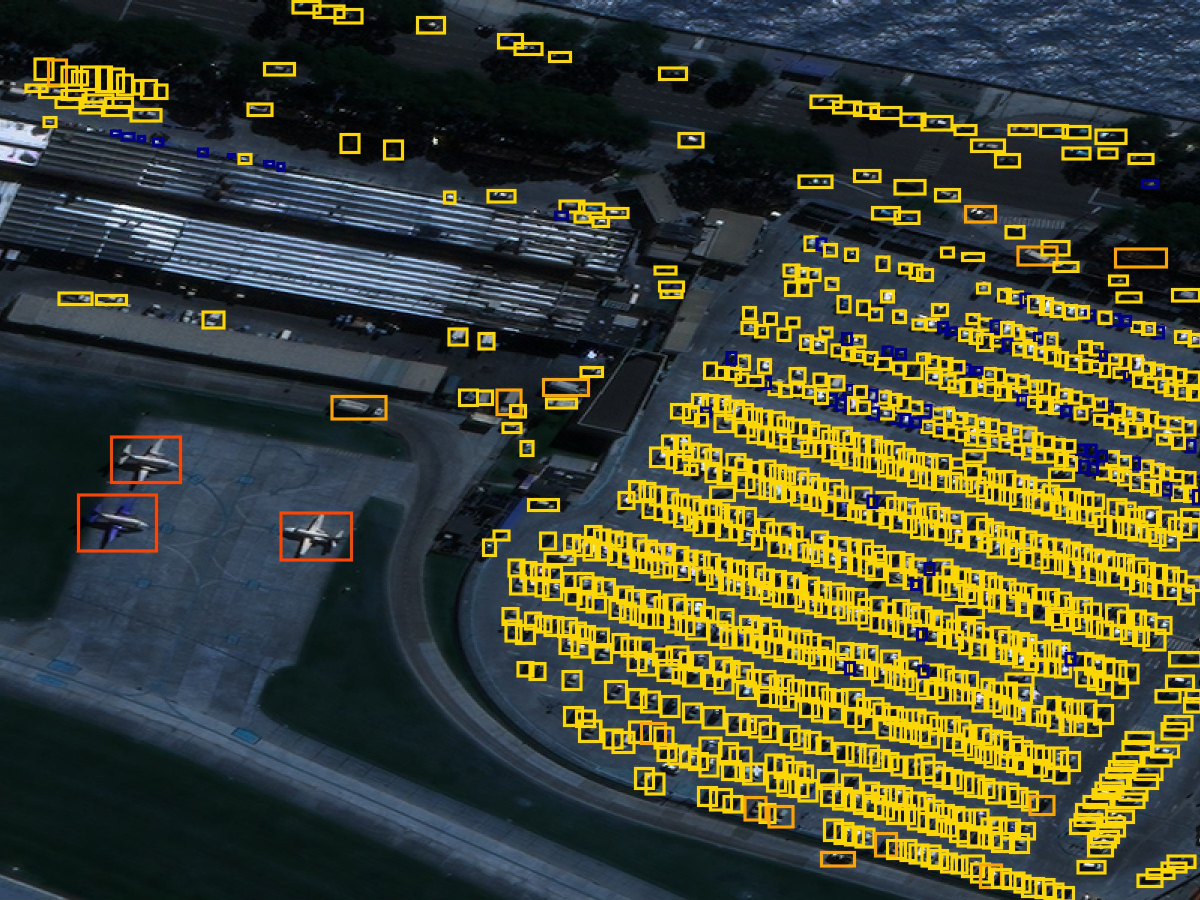}
    \caption{}
    \label{fig:rs_challenge}
\end{subfigure}
\hfill
\begin{subfigure}[b]{0.48\columnwidth}
    \centering
    \includegraphics[width=\textwidth]{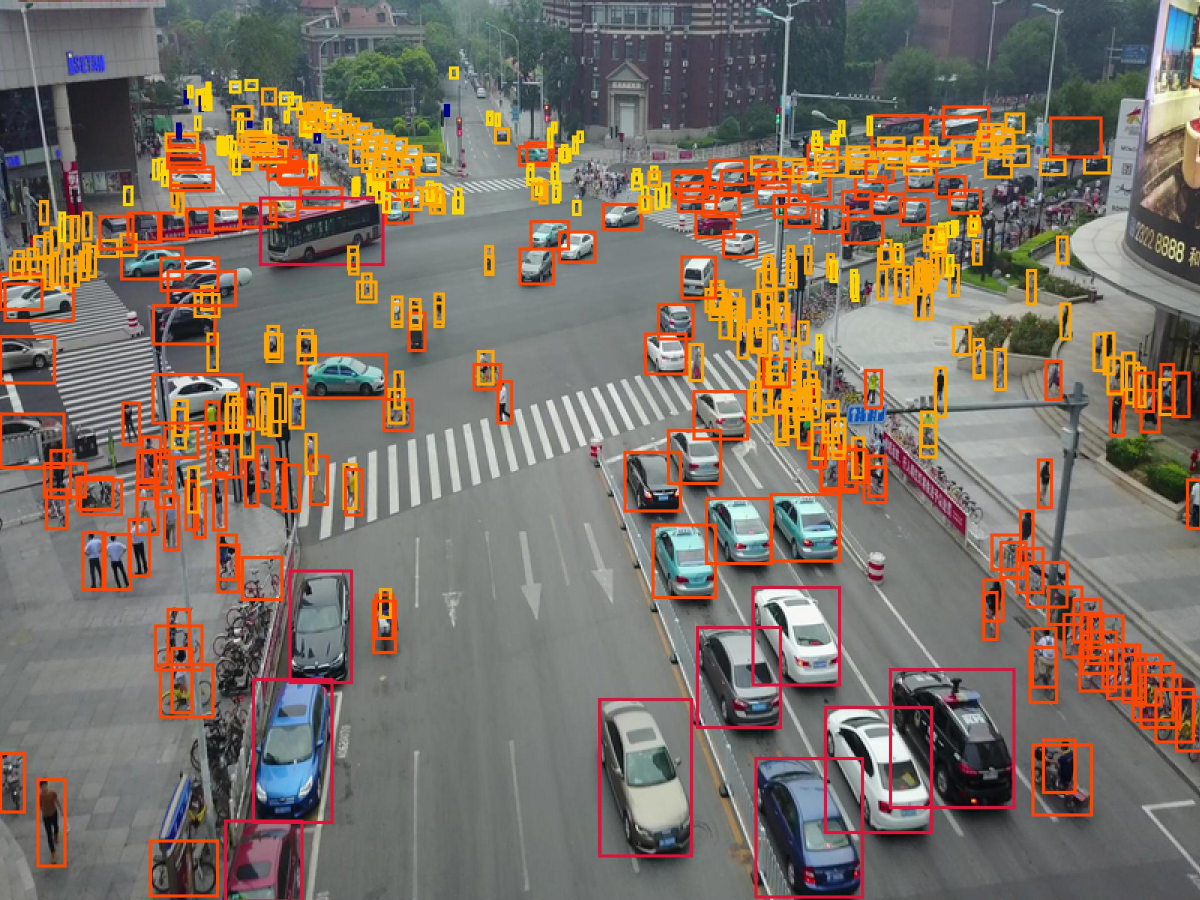}
    \caption{}
    \label{fig:aerial_challenge}
\end{subfigure}

\vspace{0.2cm}

\begin{subfigure}[b]{0.48\columnwidth}
    \centering
    \includegraphics[width=\textwidth]{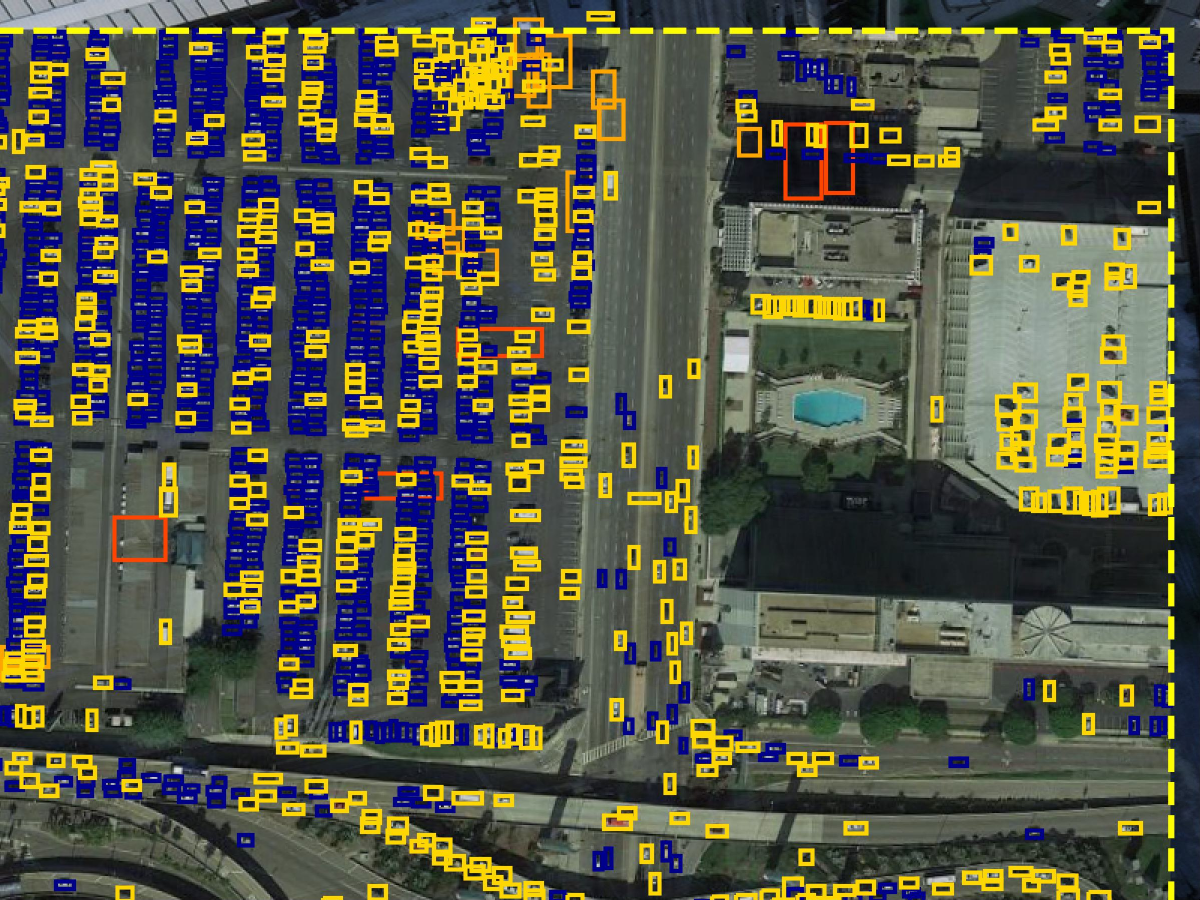}
    \caption{}
    \label{fig:scale_imbalance1}
\end{subfigure}
\hfill
\begin{subfigure}[b]{0.48\columnwidth}
    \centering
    \includegraphics[width=\textwidth]{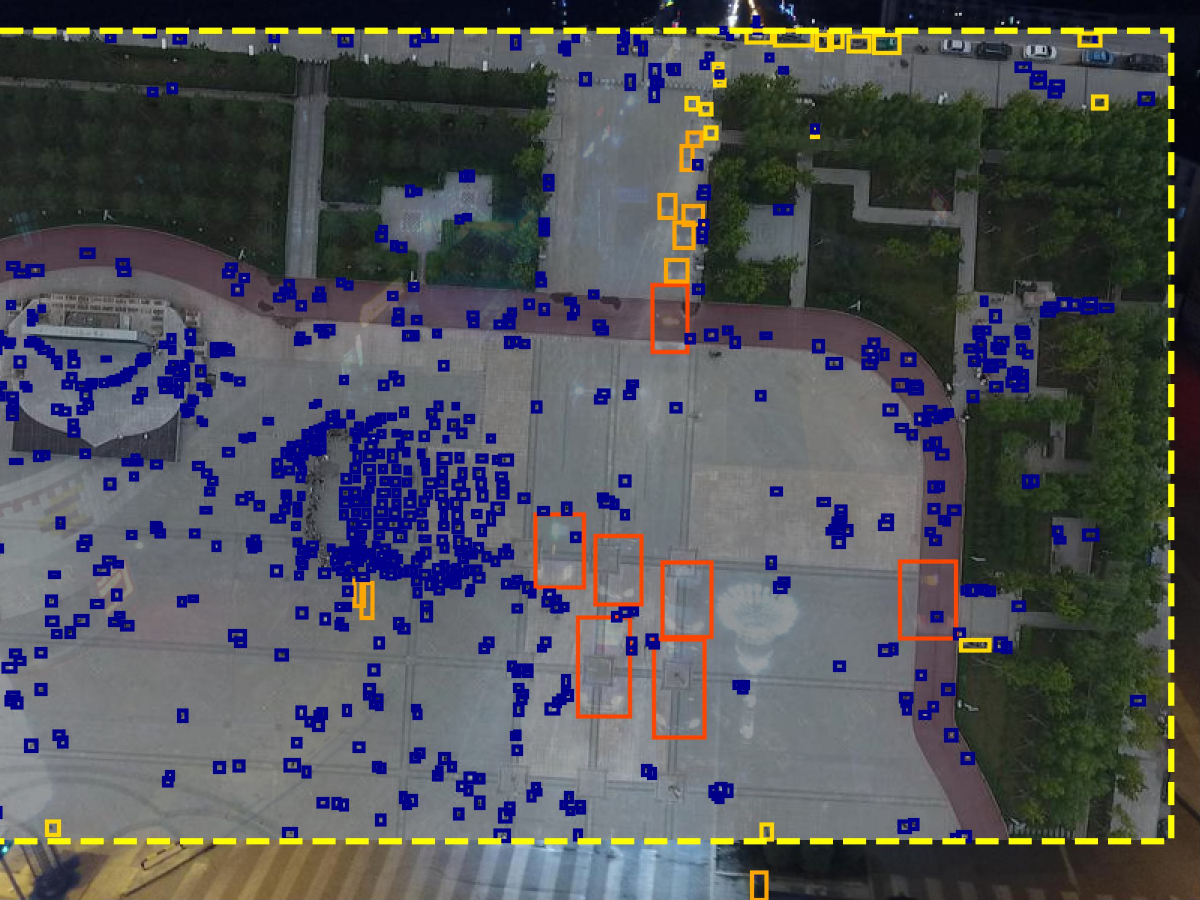}
    \caption{}
    \label{fig:scale_imbalance2}
\end{subfigure}

\caption{Multi-scale object detection challenges in remote sensing imagery. (a)-(b) Representative scenes showing the coexistence of dense tiny objects and sparse large targets. (c)-(d) Visualization of extreme scale imbalance, with dense tiny object clusters (yellow dashed regions) juxtaposed with large structures. Colors indicate scales: dark blue (very tiny $<8^2$), yellow (tiny $8^2$-$16^2$), orange (small $16^2$-$32^2$), orange-red (medium $32^2$-$96^2$), crimson (large $>96^2$).}
\label{fig:scale_distribution}
\end{figure}

\begin{figure}[!t]
\centering
\begin{subfigure}[b]{0.48\columnwidth}
    \centering
    \includegraphics[width=\textwidth]{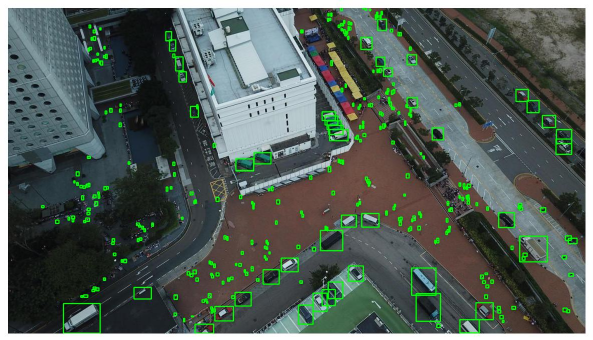}
    \caption{Ground Truth}
\end{subfigure}
\hfill
\begin{subfigure}[b]{0.48\columnwidth}
    \centering
    \includegraphics[width=\textwidth]{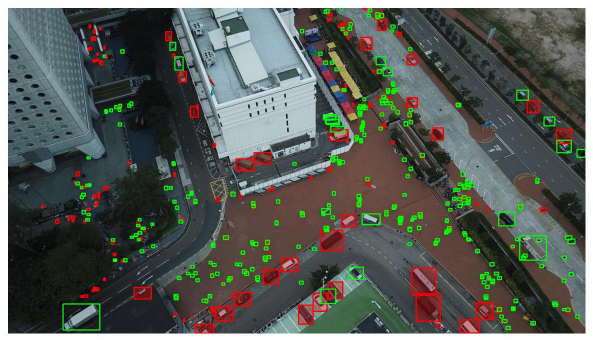}
    \caption{DQ-DETR}
\end{subfigure}

\vspace{0.2cm}

\begin{subfigure}[b]{0.48\columnwidth}
    \centering
    \includegraphics[width=\textwidth]{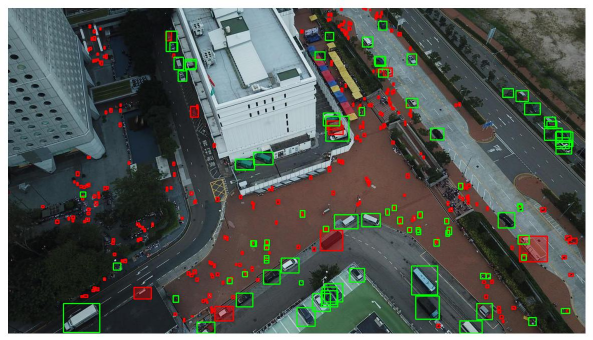}
    \caption{Ground-DINO}
\end{subfigure}
\hfill
\begin{subfigure}[b]{0.48\columnwidth}
    \centering
    \includegraphics[width=\textwidth]{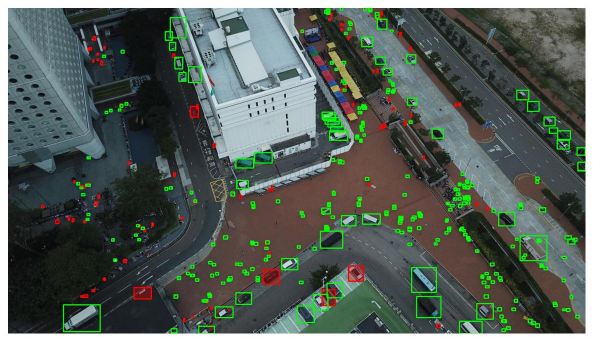}
    \caption{ScaleBridge-Det (Ours)}
\end{subfigure}

\caption{Scale balance comparative analysis with maxdet=300 constraint on a representative UAV scene containing 333 objects across five scale categories. Red boxes indicate missed detections. (a) Ground truth reference. (b) DQ-DETR excels at tiny objects but misses numerous general-scale targets. (c) Ground-DINO achieves strong general-object performance but fails on abundant tiny objects. (d) Our ScaleBridge-Det achieves balanced performance across all scales with minimal missed detections.}
\label{fig:scale_balance}
\end{figure}

% Although object detection has been significantly improved, balancing performance across all scales of objects still remains an open question, especially in remote sensing imagery \cite{8,42}. Existing detection algorithms typically focus on either small objects or large-scale objects, but suffer from poor performance segregation in scenes where the both scales are present simultaneously. When focusing on small object detection, aggressive feature enhancement and high-resolution processing are often applied to better preserve fine scales details, resulting in improvements on the mAPt (tiny-object AP)\cite{13,39}. Nevertheless, those approaches reduce the performance on medium and large objects as they are suffered from two opposite artifacts: too localized feature extraction within sub-regions making discriminant characteristics not being available at all locations while insufficient global semantic modeling results in rolled off APm and APl \cite{11,13}. As opposed to general object detectors, which have good detection performance of medium and large objects with only a few parameters, tiny targets are hard or impossible to be detected because due to the limited spatial resolution in deeper high-level layers with multi-scale feature fusion strategies \cite{33,34}. This fundamental balance between tracking tiny and locating general objects in same scenes is one of the bottlenecks that restricts to performance for real-world remote sensing applications because sparse tiny targets and regular structures often co-occur in both urban or rural landscapes \cite{8,42}.
Although object detection has achieved notable progress, balancing performance across diverse scales remains an open question, especially in remote sensing imagery \cite{zhang2024analysis,feng2023deeplearning}. Existing detection algorithms typically focus on either tiny objects or large-scale objects but suffer from poor performance segregation in scenes where both scales are present simultaneously. For tiny object detection, aggressive feature enhancement and high-resolution processing are often applied to better preserve fine-grained details, improving performance on mAPt (tiny-object AP) \cite{li2024exploring,liu2024dntr}. However, these approaches often degrade performance on medium and large objects due to two conflicting factors: overly localized features within sub-regions render discriminant characteristics unavailable at all locations, while insufficient global semantic information leads to reduced APm and APl \cite{wu2023superres,li2024exploring}. Conversely, general object detectors, which offer strong detection performance for medium and large objects with fewer parameters, find it difficult or impossible to detect tiny targets due to limited spatial resolution in the deeper layers of multi-scale feature fusion strategies \cite{ren2017fasterrcnn,lin2020retina}. This inherent conflict between detecting tiny and general objects within the same scene represents a fundamental bottleneck for real-world remote sensing applications, where tiny targets and large structures often co-occur in both urban and rural environments \cite{zhang2024analysis,feng2023deeplearning}.

\begin{figure*}[!htb]
\centering
\includegraphics[width=\textwidth]{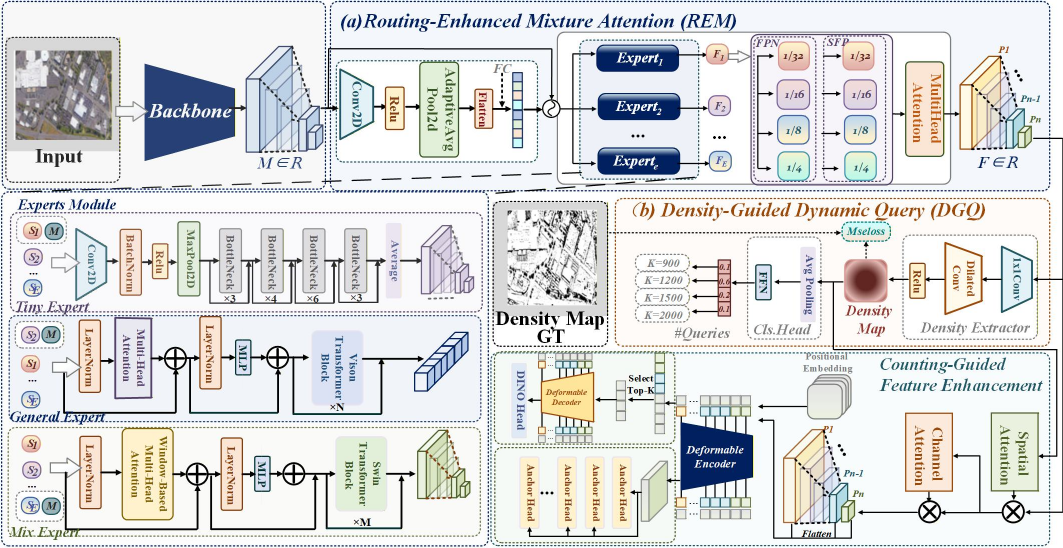}
\caption{Overview of the proposed ScaleBridge-Det framework. The framework consists of four main stages: (1) Multi-Expert Feature Extraction using diverse backbones (ResNet, ViT, Swin Transformer) with adaptive routing to select specialized experts based on input characteristics; (2) Routing-Enhanced Mixture Attention (REM) module that performs scale-adaptive feature fusion by dynamically combining expert features through hybrid attention mechanisms, generating robust multi-scale representations; (3) Density-Guided Dynamic Query (DGQ) module that predicts object density maps and adaptively adjusts query positions and numbers according to different object densities, enabling efficient resource allocation across varying density scenarios; (4) DETR Decoder with multi-layer self and cross-attention to refine predictions. The integration of these components enables balanced detection performance across extreme scale variations, from tiny objects to large structures, without performance trade-offs.}
\label{fig:framework}
\end{figure*}

Several approaches have attempted to address multi-scale detection using different techniques, but underlying limitations persist. Context-enhanced detectors such as CFENet \cite{li2024exploring} strengthen local feature representations of tiny objects by incorporating contextual enhancement modules. Nevertheless, these detectors tend to overemphasize fine-grained local features while lacking sufficient global semantic reasoning capability. Transformer-based detectors, such as DETR \cite{carion2020detr}, DINO \cite{zhang2023dino}, and CoDETR \cite{liu2023codetr}, leverage the powerful ability of transformers to capture global dependencies, improving generalization across classes. However, they still exhibit notable performance imbalance, especially for tiny objects \cite{zhao2023superres,huang2024dqdetr}. Query-based methods, such as DQ-DETR \cite{huang2024dqdetr} and D3Q \cite{guo2024d3q}, adjust query allocation according to estimated object density, theoretically facilitating better detection performance for tiny objects within dense scenes. However, these methods perform query adjustment solely based on density cues without addressing the underlying scale-completeness problem, limiting their performance when tiny and large objects occur at the same position\cite{guo2024d3q}.

Existing Large detection models have demonstrated strong potential in general object detection through self-supervised learning and massive-scale training \cite{chen2025pointmoe,chen2023adamvmoe}. DINOv3 \cite{dinov3} represents a state-of-the-art vision foundation model trained on 1.7 billion images with 7 billion parameters, achieving 66.1\% AP on COCO through self-supervised learning without manual labels. Nevertheless, such general-domain models still lack explicit mechanisms to balance detection performance across the extreme scale variations characteristic of remote sensing imagery, where tiny objects occupying fewer than 16 pixels are easily overshadowed by dominant large-object representations \cite{chen2025pointmoe,chen2023adamvmoe}. Moreover, most existing approaches rely on static feature fusion and uniform query allocation to handle all object scales. This does not sufficiently satisfy the heterogeneous feature requirements for various scale objects, making balanced detection difficult \cite{ma2023scale,aref2023transformers}.
%To address oriented object detection as well as FAFB-Net, we introduce FAMHE-Net \cite{chen2025pointmoe} which contains heterogeneous experts that focus on different scales. 

To systematically address the scale imbalance problem in remote sensing object detection, we propose ScaleBridge-Det, a unified large framework designed to achieve balanced performance across diverse object scales through scale-adaptive expert routing and density-guided query allocation. Specifically, we construct a Routing-Enhanced Mixture Attention (REM) module to flexibly combine scale-specific expert features via adaptive routing mechanisms. REM is compatible with diverse expert backbones for various scale objects and provides strong multi-scale representations for both tiny and general objects simultaneously by dynamically routing representations to the most suitable experts. Furthermore, to prevent tiny objects from being overwhelmed by the representational dominance of large targets, we propose a Density-Guided Dynamic Query (DGQ) module that adaptively adjusts query positions and numbers based on the predicted object density. DGQ predicts object density at different scales as a guide to allocate resources and adaptively manipulate query positions. This framework allows dense tiny objects to be optimized alongside general structures by integrating diverse expert backbones, such as ResNet, ViT, and SwinViT, combined with progressive training strategies. Moreover, the proposed large detection model demonstrates better cross-domain robustness in multi-source remote sensing data, as it generalizes well to different types of remote sensing imagery without extensive domain-specific fine-tuning.

% As illustrated in Figure~\ref{fig:framework} is made possible by the ability of a mixture-of-experts architecture to effectively represent both fine-grained tiny object cues and coarse objects contexts, as well in its capacity for capturing global-scale information. 

The main contributions of this work are as follows: 

%\textbf{Large detection Framework:} 
We propose ScaleBridge-Det, a first Large detection framework tailored for tiny objects that leverages a mixture-of-experts architecture and progressive training to achieve balanced performance across extreme scale variations in remote sensing imagery. This addresses the fundamental scale competition problem where tiny objects are suppressed by dominant large structures.

%\textbf{Scale-Adaptive Expert Routing:} 
We develop a Routing-Enhanced Mixture Attention (REM) module that dynamically selects and fuses scale-specific expert features through adaptive routing mechanisms. This generates robust multi-scale representations that prevent performance trade-offs between tiny and general object detection.

%\textbf{Density-Guided Query Allocation:} 
We propose a Density-Guided Dynamic Query (DGQ) module that predicts object density and adaptively adjusts query positions and numbers. This improves APvt and APt by 1.9\% through scale-aware resource allocation and mitigates out-of-memory issues in dense scenes.

%\textbf{State-of-the-Art Performance:} 
Extensive experiments demonstrate that ScaleBridge-Det achieves state-of-the-art performance with 35.7\% mAP on AI-TODv2, 29.8\% mAP on DTOD and 30.7\% mAP on VisDrone. It maintains balanced detection accuracy across all object scales while exhibiting superior cross-domain robustness as an additional benefit of the large model capacity.

The remainder of this paper is organized as follows: Section II reviews related work in object detection for remote sensing. Section III provides a detailed description of the proposed ScaleBridge-Det framework. Section IV presents comprehensive experimental results and analysis. Section V concludes the paper with discussions and future directions.

% \begin{figure*}[!t]
% \centering
% \includegraphics[width=0.95\textwidth]{bibtex/fig/detection_comparison_insets1.png}
% \caption{Motivating example: Detection comparison on an extremely dense scene (4766×2735 pixels, 3058 objects). The main image shows Ground Truth annotations (green boxes) with a highlighted zoom region (golden dashed box) containing 890 densely packed objects. Two zoomed insets compare state-of-the-art DINO-X (top-right, red boxes: 108 detections, 12\% recall) versus our ScaleBridge-Det (bottom-right, blue boxes: 659 detections, 74\% recall). Light green boxes in insets show Ground Truth for reference. DINO-X exhibits catastrophic performance degradation with massive missed detections, while our method maintains robust detection capability through density-guided query allocation and scale-adaptive expert routing. This example demonstrates the critical challenge motivating our work: achieving balanced detection performance under extreme object density and scale variation.}
% \label{fig:dense_detection_comparison}
% \end{figure*}

 \section{Related Work}
This section reviews the literature most relevant to this paper, organized into three categories: general and remote sensing object detection, tiny object detection methods, and Large detection models.

\subsection{General Object Detection}

\IEEEPARstart{O}{bject} detection in remote sensing imagery faces fundamental challenges that distinguish it from general computer vision tasks. Traditional CNN-based detectors such as Faster R-CNN \cite{ren2017fasterrcnn} and RetinaNet \cite{lin2020retina} achieve strong performance on natural images but struggle when confronted with the extreme scale variations and dense distributions characteristic of aerial imagery \cite{li2023smallobj,cheng2023small}. These methods fail to address how detectors can maintain balanced performance when tiny objects occupying fewer than 32$\times$32 pixels coexist with large structures spanning hundreds of pixels within the same scene.

Transformer-based architectures have emerged as a foundational approach for capturing long-range dependencies in complex visual scenes. DETR \cite{carion2020detr} reformulates detection as a direct set prediction problem, eliminating hand-crafted components through end-to-end training. However, the method exhibits slow convergence and limited performance on small objects, achieving only 42\% AP on COCO. Deformable DETR \cite{zhu2021deformabledetr} improves convergence through deformable attention mechanisms that sample features at sparse key locations, yet the fundamental limitation persists. It still struggles to efficiently allocate attention across objects with extreme scale variations while maintaining balanced performance across all object scales.

DINO \cite{zhang2023dino} advances DETR-like models by developing contrastive denoising training and mixed query selection, achieving 51.3\% AP on COCO. CoDETR \cite{liu2023codetr} proposes collaborative hybrid assignment training to enhance encoder-decoder optimization, reaching 64.4\% AP. Despite these improvements, these methods face critical challenges in dense remote sensing scenarios \cite{zhao2023superres,huang2024dqdetr}. The core problem is that existing architectures lack mechanisms to prevent large objects from suppressing tiny object detection during training and inference.

Recent DETR variants, including DAB-DETR \cite{liu2022dabdetr}, DN-DETR \cite{li2022dndetr}, and Conditional DETR \cite{meng2021conditionaldetr}, develop dynamic anchors and conditional queries to improve convergence. Group DETR \cite{chen2022groupdetr} introduces one-to-many assignment strategies for training efficiency. While these methods demonstrate potential for capturing global context, they require substantial adaptation for remote sensing applications where the scale gap between tiny and large objects creates fundamental optimization conflicts \cite{li2023maskdino,zhang2023dense}.

\subsection{Tiny Object Detection}
Tiny object detection in remote sensing presents distinctive challenges due to limited pixel representation, complex background clutter, and heterogeneous object densities \cite{li2023smallobj,cheng2023small}. The core difficulty lies in extracting sufficiently discriminative features from objects occupying minimal spatial extent while maintaining computational efficiency, a constraint that fundamentally distinguishes this task from general object detection.

Context-enhanced approaches attempt to address feature representation challenges. CFENet \cite{li2024exploring} develops global and local context enhancement modules to improve small object features, achieving 27.8\% APt on AI-TOD. However, overemphasis on local details leads to insufficient global semantic information, reducing APm performance by 4.2\% \cite{li2024exploring}. This reveals a critical trade-off that methods optimized for tiny objects often sacrifice performance on larger ones. FPN-based methods \cite{lin2017fpn} employ multi-scale feature pyramids to handle scale variations, but static allocation strategies fail when target densities vary dramatically across scenes \cite{wang2024query,ma2023scale}.

Feature enhancement techniques have been extensively explored. QueryDet \cite{chen2022querydet} proposes adaptive query refinement for small objects, while TPH-YOLOv5 \cite{zhu2021tphyolov5} develops transformer prediction heads to capture long-range dependencies in UAV images. DNTR \cite{liu2024dntr} employs denoising feature pyramids to filter background noise, achieving improvements in APt and APvt on aerial datasets. Scale-aware attention networks \cite{dai2021dynamicdetr} dynamically adjust receptive fields based on object scale, and SAHI \cite{akyon2022sahi} performs detection on sliced images to increase effective resolution. Despite these advances, these methods address symptoms rather than the root cause of scale imbalance, lacking mechanisms to explicitly balance performance across all object scales.

Density-aware methods tackle the problem of resource allocation in varying density scenarios. Fixed query mechanisms in DETR-like methods create inefficiency when object density varies dramatically—from sparse scenes with few objects to dense scenarios with hundreds of tiny targets \cite{feng2023deeplearning}. DQ-DETR \cite{li2024exploring} develops categorical counting modules and density-guided feature enhancement, achieving 30.2\% mAP on AI-TODv2 with 42.1\% improvement in APvt for dense images. D3Q \cite{guo2024d3q} refines this approach by explicitly estimating object density to determine optimal query numbers, reaching 32.1\% mAP on AI-TODv2. Other approaches including Dense Query DETR \cite{sun2021sparsedetr}, Adaptive Query Selection \cite{wang2023adaptive}, and QueryProp \cite{wang2023queryprop} explore dynamic query strategies.

Despite progress, density estimation remains challenging in complex scenes with occlusion and varying illumination \cite{zhang2024analysis,wang2023smallobject}. More critically, existing approaches restrict their scope to density-based query adjustment while neglecting heterogeneous feature requirements across different scales \cite{liu2023domain,aref2023transformers}. Existing detection frameworks cannot simultaneously allocate computational resources and representational capacity adaptively to prevent large objects from suppressing tiny target detection.

\subsection{Large Detection Models}

Large detection models leverage increased model capacity to address heterogeneous feature requirements emerging from diverse data distributions and complex detection scenarios. However, achieving balanced performance across extreme scale variations while maintaining computational efficiency remains a fundamental challenge. This requires architectures capable of adaptively allocating representational capacity without incurring prohibitive computational overhead.

Large vision foundation models have achieved remarkable success through self-supervised learning at scale. DINOv3 \cite{dinov3} demonstrates that large-scale pre-training on natural images can achieve strong performance on COCO object detection with frozen backbones. However, these general-domain models struggle with cross-domain generalization to remote sensing scenarios, where object scales and densities fundamentally differ from natural images \cite{chen2025pointmoe}. To address this domain gap, LAE-DINO \cite{pan2025locate} introduces the first large-scale open-vocabulary detector specifically designed for remote sensing, incorporating Dynamic Vocabulary Construction (DVC) and Visual-Guided Text Prompt Learning (VisGT) modules. Trained on the LAE-1M dataset with 180M parameters including a BERT-based language model, LAE-DINO achieves strong performance on remote sensing benchmarks. FAMHE-Net \cite{chen2025pointmoe} further introduces multi-scale feature augmentation with heterogeneous experts. While these methods demonstrate that diverse representations can improve detection, they lack explicit mechanisms to balance performance across the extreme scale variations characteristic of aerial imagery.

Sparse expert architectures provide computational efficiency while scaling model capacity. Mixture-of-Experts (MoE) implementations activate subsets of parameters based on input characteristics, enabling large models with manageable computational costs \cite{chen2023visionmoe,lin2024moellava}. Multi-task MoE variants develop adaptive expert selection strategies \cite{chen2023adamvmoe,ma2018mmoe}, while recent work explores expert architectures for 3D tasks \cite{chen2025pointmoe2}.

Despite success in generic vision tasks, applying large models to tiny object detection in remote sensing remains under-explored. Existing approaches struggle to handle density imbalances and scale competition where large objects suppress tiny targets \cite{zhang2024analysis,feng2023deeplearning}. Routing strategies in current architectures do not explicitly consider spatial feature variations across remote sensing modalities (satellite, aerial, UAV), limiting cross-domain robustness \cite{liu2023domain,dosovitskiy2021vit}. Recent work, including Sparse Expert networks \cite{zhou2022moeexpert} and Hierarchical MoE \cite{li2024hierarchical}, begins addressing dynamic expert activation and hierarchical feature processing. However, there is a critical gap between the comprehensive integration of large architectures with scale-adaptive routing and density-aware mechanisms for balanced tiny-to-large object detection. This gap motivates our proposed ScaleBridge-Det framework, which addresses the fundamental question of how large detection models can achieve balanced performance across extreme scale variations through adaptive expert routing and density-guided resource allocation.

\begin{figure*}[!t]
\centering
\includegraphics[width=\textwidth]{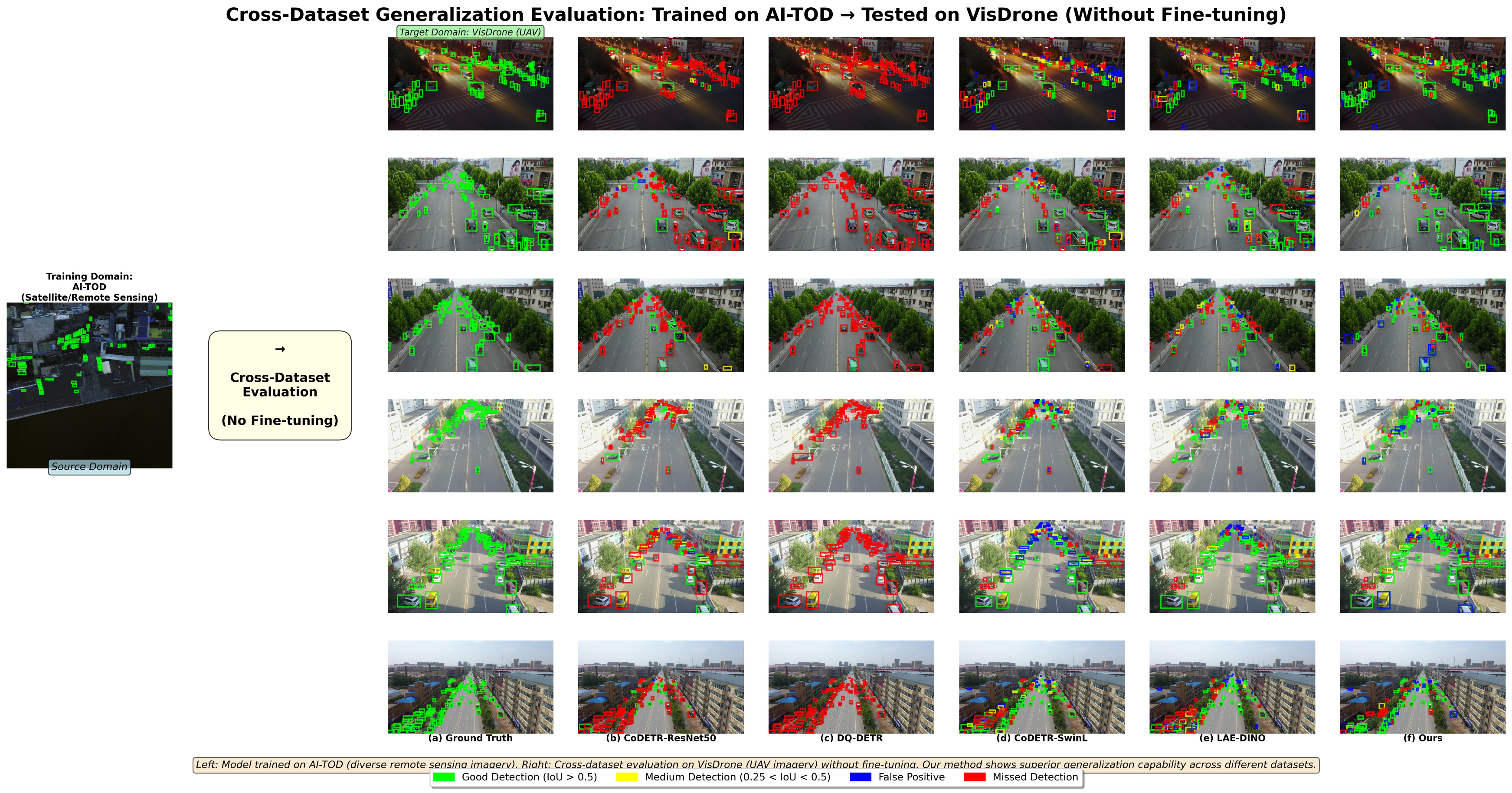}
\caption{Comprehensive cross-domain evaluation visualization with category mapping. \textbf{Left}: Training domain showing AI-TOD satellite/remote sensing imagery (100 objects). \textbf{Middle}: Cross-domain transfer with category mapping (AI-TOD vehicle/person mapped to corresponding VisDrone categories) and no fine-tuning. \textbf{Right}: Testing domain with 6 VisDrone UAV test images, each showing 6 model comparisons. Detection quality is color-coded: \textcolor{green}{Green} boxes indicate good detections (IoU $>$ 0.5), \textcolor{yellow}{Yellow} boxes show medium detections (0.25 $<$ IoU $<$ 0.5), \textcolor{blue}{Blue} boxes represent false positives, and \textcolor{red}{Red} boxes mark missed detections. Column labels: (a) Ground Truth, (b) Faster R-CNN, (c) DETR, (d) DQ-DETR, (e) CoDETR, (f) ScaleBridge-Det (Ours, highlighted with red border). Our method demonstrates superior cross-domain generalization with significantly more green boxes (good detections) and fewer red boxes (missed detections) compared to baseline methods, validating the effectiveness of scale-adaptive expert routing and density-guided query allocation for cross-domain transfer.}
\label{fig:cross_domain_comprehensive}
\end{figure*}

\section{Method}
% 1.文中要明确说fig:framework的backbone部分和encoder、decoder部分为baseline原有的
% 2.调整行文顺序，先介绍完整的专家系统和模块设计，最后介绍训练策略
% 3.关于动态激活值，明确说明 为了充分利用每一类专家的特征特性，动态调整激活值(最低优先)保证专家特征完整和推理效率
% 4.关于self-attention，图表中需要明确注明是如何使用，以增强可读性与行文一致性

This section first describes the framework of the proposed ScaleBridge-Det. Subsequently, the REM and DGQ modules are elaborated. Finally, the training strategy is introduced.

\subsection{Overall Framework}
The overall framework of the proposed ScaleBridge-Det is depicted in Fig. 3, which builds upon the CoDETR baseline \cite{liu2023codetr}. ScaleBridge-Det consists of four key components: a backbone network, a REM module, a DGQ module, and a DETR decoder. Algorithm~\ref{alg:scalebridge} presents the detailed pseudo-code of the complete ScaleBridge-Det framework. The algorithm takes an image tensor $\mathbf{I} \in \mathbb{R}^{B \times C \times H \times W}$ as input, where $B$ is the batch size, $C$ is the number of channels (typically 3 for RGB), $H$ is the height, and $W$ is the width. $\mathbf{I}$ is fed into the backbone for feature extraction. Furthermore, REM employs diverse backbone architectures, including ResNet, ViT, and Swin Transformer, to capture scale-specific representations. Scale-adaptive feature fusion is adopted to dynamically combine expert features through hybrid attention mechanisms. Then, the DGQ module adaptively adjusts query allocation based on predicted object density, incorporating a counting-guided feature enhancement mechanism that utilizes channel and spatial attention (CBAM) to refine encoder features according to density predictions, enabling more accurate query positioning in dense regions. Finally, the DETR decoder inherited from the baseline applies multi-layer self- and cross-attention, culminating in prediction heads for final bounding box regression and classification.
%This design enables the model to handle objects of varying scales and adapt to remote sensing images from different sources and resolutions \cite{fedus2022switch,shazeer2017moe}, while the density-guided mechanism provides spatial priors on object distribution for accurate localization in complex backgrounds \cite{guo2024d3q}.

%The algorithm proceeds through multi-expert feature extraction with adaptive routing to select specialized backbone networks, followed by the REM module for scale-adaptive feature fusion using hybrid attention. The DGQ module then generates density maps and performs dynamic query allocation. Subsequently, the DETR decoder inherited from the baseline applies multi-layer self and cross-attention, culminating in prediction heads for final bounding box regression and classification. The key innovation lies in the integration of routing-based expert selection with density-guided query adaptation, enabling balanced detection across extreme scale variations.

\subsection{Routing-Enhanced Mixture Attention (REM)}
The Routing-Enhanced Mixture Attention (REM) module addresses the fundamental challenge of integrating diverse expert features for multi-scale object detection. This module tightly couples routing-based expert selection with hybrid attention mechanisms for multi-scale feature fusion, ensuring that dynamic expert routing directly informs feature pyramid construction and enhances the model's capacity to handle diverse scales and feature characteristics \cite{lin2017fpn}.

\textbf{Adaptive Expert Routing:} As illustrated in Fig.~\ref{fig:framework}(a), the REM module employs an input-dependent dynamic routing mechanism that adaptively selects specialized experts based on image-specific characteristics \cite{shazeer2017moe,zhou2022moeexpert}. Given feature representations $M$ from the base backbone (denoted as "$M \in R$" in Fig.~\ref{fig:framework}), the routing network processes features through Conv2d, ReLU, AdaptiveAvgPool2d, Flatten, and FC layers to compute expert routing scores:
\begin{equation}
[s_1, s_2, \ldots, s_E] = \text{Softmax}(\text{FC}(\text{Flatten}(\text{GAP}(M))))
\end{equation}
where $s_e \in [0,1]$ denotes the routing weight for expert $e$, indicating its importance for the current input.

To enhance computational efficiency and prevent feature dilution from low-contribution experts, we employ a selective activation mechanism that ensures at least one expert from each category is activated:
\begin{equation}
\mathcal{A} = \{e \mid e \in \text{Top-1}(\mathcal{C}_e)\}
\end{equation}
where $\mathcal{A}$ denotes the set of activated experts, and $\mathcal{C}_e$ represents the category of expert $e$ (tiny/general/mix). This category-wise top-1 selection guarantees diversity across expert types while maintaining computational efficiency. As shown in the \enquote{Experts Module} of Fig.~\ref{fig:framework}, the activated heterogeneous experts produce feature outputs $F_e$ for $e \in \mathcal{A}$, which are then weighted by the corresponding routing scores $s_e$ for adaptive fusion.

\textbf{Scale-Adaptive Feature Fusion:} Following expert routing, each activated expert $e \in \mathcal{A}$ extracts multi-scale feature pyramids through FPN or SFP, producing features $F_e^l$ at each pyramid level $l$.

To address the magnitude imbalance across heterogeneous expert architectures, we introduce a scale-aware normalization mechanism:
\begin{equation}
\widetilde{F}_e^l = \alpha_e \cdot F_e^l, \quad \forall e \in \mathcal{A}
\end{equation}
where $\alpha_e$ is the learnable expert-specific calibration factor optimized end-to-end during training.

Building upon normalized features, a learnable dynamic gating mechanism computes scale-specific fusion weights. At each pyramid level $l$, the gating network aggregates expert features through concatenation and global average pooling, then computes level-specific gating scores:
\begin{equation}
g_l=\text{GAP}(\text{Concat}(\{\widetilde{F}_e^l \mid e \in \mathcal{A}\}))
\end{equation}
\begin{equation}
\quad w_l=\text{Softmax}(W_l \cdot g_l + b_l)
\end{equation}
where $W_l$ and $b_l$ are scale-specific learnable parameters. The final fused representation at each scale combines both routing scores $s_e$ and level-specific weights $w_{l,e}$ for activated experts only:
\begin{equation}
F_l^{\text{fused}} = \sum_{e \in \mathcal{A}} s_e \cdot w_{l,e} \cdot \widetilde{F}_e^l
\end{equation}

The fused multi-scale features are subsequently processed through the Multi-Head Attention mechanism (shown in Fig.~\ref{fig:framework}), which performs hierarchical feature integration using self-attention:
\begin{equation}
F^{\text{final}} = \text{MultiHead}(F^{\text{fused}}) = \text{Softmax}\left(\frac{QK^T}{\sqrt{d}}\right)V
\end{equation}
where $Q$, $K$, $V$ are linear projections of the fused features. This generates the output feature pyramids $\{P_1, \ldots, P_n\}$ that serve as inputs to the downstream DETR decoder, balancing local details and global context across scales.

\subsection{Density-Guided Dynamic Query (DGQ)}

The Density-Guided Dynamic Query (DGQ) module enhances the detection of dense tiny object regions by providing spatial priors on object distribution \cite{guo2024d3q}. Integrated with the highest-resolution feature map (256$\times$256) from REM, the DGQ module combines density map generation with instance count-aware dynamic query allocation, significantly improving detection performance in high-density scenarios. Traditional fixed-query mechanisms prove insufficient for tiny object detection in remote sensing, where objects often cluster densely with limited pixel information \cite{huang2024dqdetr,guo2024d3q}. The density map provides spatial priors that mitigate query starvation in dense regions and improve recall for small-scale targets \cite{wang2024query}. Furthermore, in images containing thousands of targets, fixed-query approaches encounter out-of-memory (OOM) errors during Hungarian matching due to the quadratic complexity of the query count. The proposed dynamic adjustment mechanism scales queries appropriately to maintain both efficiency and performance. The detailed algorithm pipeline is presented in Algorithm~\ref{alg:scalebridge}.

\textbf{Density Map Generation:} The ground truth density map is generated by convolving object centers with a Gaussian kernel, capturing local object densities. For each object instance centered at position $c_i = (x_i, y_i)$, a 3$\times$3 Gaussian kernel with standard deviation $\sigma = 1.5$ (empirically optimized for tiny objects) is applied to produce the density map $D^{gt}$:
\begin{equation}
D^{gt}(p) = \sum_{i} \exp\left( -\frac{\|p - c_i\|^2}{2\sigma^2} \right),
\end{equation}
where $p$ denotes pixel positions, and the map is normalized to reflect relative densities. During training, a lightweight convolutional head (comprising 3$\times$3 convolutions followed by 1$\times$1 convolution) processes the high-resolution feature map to predict the density map $D^{pred}$. The module is supervised using mean squared error (MSE) loss for density map consistency, combined with auxiliary classification loss for semantic alignment:
\begin{equation}
\mathcal{L}_{density} = \| D^{pred} - D^{gt} \|_2^2 + \lambda_{cls} \mathcal{L}_{cls}(D^{pred}),
\end{equation}
where $\mathcal{L}_{cls}$ is the binary cross-entropy loss encouraging high density predictions in object regions. This dual supervision ensures the predicted density map accurately approximates the ground truth distribution while effectively discriminating foreground from background \cite{guo2024d3q}.

The predicted density map serves as a spatial prior, highlighting regions with high object concentration. This approach draws inspiration from crowd counting techniques adapted for detection tasks, where density maps improve localization in cluttered scenes by providing coarse estimates of object counts and positions before fine-grained detection \cite{zhu2021deformabledetr}. For tiny objects in aerial images, the density map compensates for detailed feature loss during downsampling in deeper layers, preserving cluster information that might otherwise be diluted.

\textbf{Dynamic Query Selection:} To adapt to varying object densities, the total instance count $N$ is estimated by integrating over the predicted density map:
\begin{equation}
N = \sum D^{pred},
\end{equation}
rounded to the nearest integer. Based on this estimate, query numbers are dynamically adjusted across four tiers to balance computational efficiency and detection capacity:

\begin{equation*}
\text{Queries} =
\begin{cases}
900  & \text{if } N \leq 10, \\
1200 & \text{if } 10 < N \leq 100, \\
1500 & \text{if } 100 < N \leq 500, \\
2000 & \text{if } N > 500.
\end{cases}
\end{equation*}

This tiered strategy prevents under-querying in dense scenarios (leading to missed detections) and over-querying in sparse scenes (increasing noise and slowing convergence). To enhance query initialization, initial query positions are sampled from high-density regions in $D^{pred}$, using weighted sampling where the selection probability is proportional to density values. This density-oriented initialization, following principles from recent DETR variants \cite{zhang2023dino,guo2024d3q}, focuses queries on potential object clusters from the outset, improving transformer attention allocation for tiny, dense targets.

The DGQ module synergizes with REM by leveraging rich multi-scale features for accurate density prediction, while dynamic queries feed into the DETR decoder to refine bounding box predictions. Compared to prior density-guided methods \cite{huang2024dqdetr,guo2024d3q}, integration within the large framework offers superior cross-scale adaptability and noise robustness, as experts can specialize in density estimation under varying conditions. This module addresses key DETR limitations for tiny object detection, including slow convergence on small scales and poor handling of density variations, making it essential for real-world remote sensing applications.

\subsection{Progressive Training Strategy}

Training large multi-expert detection models presents significant challenges due to parameter explosion, training instability, and gradient propagation issues inherent in directly combining multiple heterogeneous experts. To address these challenges, a progressive joint training paradigm is adopted through structured expert integration, where expert backbones undergo individual pre-training to establish stable and effective feature representations on independent tasks before integration into the complete framework.

General vision experts receive pre-training on COCO and LVIS datasets to acquire broad visual feature learning capabilities, while remote sensing-specific experts are pre-trained on DOTA and DIOR datasets to capture domain-specific aerial imagery characteristics. This domain-specialized pre-training ensures that each expert develops strong foundational representations aligned with its designated architectural paradigm and task requirements. The individual pre-training stage proves particularly critical for maintaining expert diversity, as it prevents premature feature homogenization that commonly occurs when heterogeneous architectures are trained jointly from initialization.

Expert integration proceeds through gradual introduction in a structured order, beginning with smaller expert combinations and progressively incorporating additional experts while conducting parameter alignment and collaborative optimization at each integration stage \cite{fedus2022switch}. This gradual integration strategy alleviates convergence difficulties common to large expert systems while promoting complementary feature learning among experts with distinct architectural inductive biases. During integration, expert backbone parameters remain frozen to preserve their specialized feature extraction capabilities, while the routing network, gating mechanisms, and detection heads undergo joint optimization. This selective parameter updating strategy significantly reduces GPU memory requirements by eliminating gradient computation for expert backbones, enabling efficient training despite the large model capacity.

The complete detection system emerges through the selection of the best-performing expert combinations based on validation performance across multiple criteria, including overall detection accuracy,  scale-specific AP metrics, and cross-domain generalization capability. The entire framework, encompassing REM and DGQ modules alongside selected expert ensembles, undergoes joint end-to-end fine-tuning on target remote sensing datasets to optimize collaborative feature fusion and query allocation strategies. This progressive approach substantially reduces training instability while maintaining the representational benefits of diverse expert architectures, providing a robust foundation for Large detection model construction. To balance model performance with computational efficiency during deployment, only the most effective expert combinations identified through validation are retained in the final model architecture.

\begin{algorithm}[!t]
\small
\caption{ScaleBridge-Det: Complete Detection Pipeline}
\label{alg:scalebridge}
\KwIn{Image $\mathbf{I} \in \mathbb{R}^{B \times 3 \times H \times W}$}
\KwOut{Detections $\mathcal{R} = \{(\text{bbox}_i, \text{score}_i, \text{class}_i)\}$}

\tcp{Stage 1: Adaptive Expert Routing}
$M \gets \text{BaseBackbone}(\mathbf{I})$ \tcp*{Extract base features}
$[s_1, s_2, \ldots, s_E] \gets \text{Softmax}(\text{FC}(\text{Flatten}(\text{GAP}(M))))$
$\mathcal{A} \gets \{e \mid e \in \text{Top-1}(\mathcal{C}_e)\}$ \tcp*{Select top-1 per category}

\tcp{Stage 2: Expert Feature Extraction}
\ForEach{expert $e \in \mathcal{A}$}{
    $F_e \gets \text{Expert}_e(\mathbf{I})$ with frozen parameters\;
    $\{F_e^1, F_e^2, \ldots, F_e^L\} \gets \text{FPN/SFP}(F_e)$
}

\tcp{Stage 3: Scale-Adaptive Feature Fusion}
\ForEach{pyramid level $l \in \{1, \ldots, L\}$}{
    $\widetilde{F}_e^l \gets \alpha_e \cdot F_e^l$ for each $e \in \mathcal{A}$\;

    $g_l \gets \text{GAP}(\text{Concat}(\{\widetilde{F}_e^l \mid e \in \mathcal{A}\}))$\;
    $w_l \gets \text{Softmax}(W_l \cdot g_l + b_l)$\;

    $F_l^{\text{fused}} \gets \sum_{e \in \mathcal{A}} s_e \cdot w_{l,e} \cdot \widetilde{F}_e^l$\;
}

\tcp{Multi-head attention for final features}
$F^{\text{final}} \gets \text{MultiHead}(F^{\text{fused}}) = \text{Softmax}(\frac{QK^T}{\sqrt{d}})V$\;

\tcp{Stage 4: Density-Guided Dynamic Query}
$D^{pred} \gets \text{DensityHead}(F^{\text{final}})$ \tcp*{Predict density map}
$N_q \gets \text{round}(\sum D^{pred})$ \tcp*{Estimate object count}
% $N_q \gets \begin{cases}
% 900 & \text{if } N \leq 10\\
% 1200 & \text{if } 10 < N \leq 100\\
% 1500 & \text{if } 100 < N \leq 500\\
% 2000 & \text{if } N > 500
% \end{cases}$ \tcp*{Adaptive query allocation}
$Q_{\text{init}} \gets \text{WeightedSample}(D^{pred}, N_q)$ \tcp*{Sample from density map}

\tcp{Stage 5: DETR Decoder with Dynamic Queries}
$Q \gets Q_{\text{init}}$\;
\For{decoder layer $i \gets 1$ \KwTo $N_{\text{layers}}$}{
    $Q \gets \text{SelfAttention}(Q)$\;
    $Q \gets \text{CrossAttention}(Q, F^{\text{final}})$\;
    $Q \gets \text{FFN}(Q)$\;
}

\tcp{Stage 6: Prediction Heads}
$\{\text{bbox}, \text{class}\} \gets \text{PredictionHeads}(Q)$\;
$\mathcal{R} \gets \{(\text{bbox}_i, \text{score}_i, \text{class}_i) \mid \text{score}_i > \tau_{\text{conf}}\}$\;

\Return{$\mathcal{R}$}
\end{algorithm}

\section{Experiments}
In this section, we first describe the datasets and evaluation metrics used in our experiments. Subsequently, we provide implementation details of the proposed method. Then, we compare ScaleBridge-Det with state-of-the-art approaches on multiple benchmarks. Finally, we validate the effectiveness of key components through comprehensive ablation studies.

\subsection{Datasets and Evaluation Metrics}

\textbf{AI-TOD Dataset:} The AI-TODv2 dataset comprises 28,036 aerial images containing 700,621 object instances, with 86\% of objects smaller than 16 pixels and an average object size of only 12.8 pixels \cite{aitod}. This extreme tiny-object emphasis makes AI-TOD a particularly challenging benchmark for evaluating balanced detection performance across scales. The dataset uses a standard split of 11,414 images for training, 2,804 for validation, and 14,018 for testing.

\textbf{VisDrone Dataset:} VisDrone contains 10,209 UAV-captured images with 2000$\times$1500 pixel resolution, covering 10 object categories with splits of 6,471 training, 548 validation, and 3,190 test images \cite{visdrone}. For cross-domain evaluation, we establish a category mapping protocol to align semantically compatible classes between AI-TOD and VisDrone. Specifically, AI-TOD \textit{vehicle} (category 5) is mapped to VisDrone vehicle-related categories \{\textit{car, van, truck, tricycle, awning-tricycle, bus, motor}\}, and AI-TOD \textit{person} (category 6) is mapped to VisDrone human-related categories \{\textit{pedestrian, people}\}. Other AI-TOD categories (airplane, bridge, storage-tank, ship, swimming-pool, wind-mill) have no correspondence in VisDrone and are excluded from cross-domain evaluation. Models trained on AI-TOD are directly evaluated on the VisDrone validation set without fine-tuning, providing a rigorous test of cross-domain generalization under distribution shifts between satellite and UAV imaging platforms.

\textbf{DTOD Dataset:} The DTOD dataset presents an orthogonal challenge through extreme object density, with individual images containing over 2,000 object instances on average \cite{dtod}. This dataset complements our evaluation by testing detection performance in scenarios with overwhelming object crowding, where density-guided query allocation becomes particularly critical for maintaining computational efficiency and detection accuracy.

\textbf{Evaluation Metrics:} We evaluate model performance using standard metrics including Average Precision (AP), AP50 (IoU threshold 0.5), and scale-specific metrics: APvt (very tiny), APt (tiny), APs (small), and APm (medium). For cross-domain evaluation, we report mAP and mAP50 to assess generalization capability.

\subsection{Implementation Details}

Our implementation builds upon the MMDetection framework. All models are trained with uniform input resolution of 1024$\times$1024 across all datasets (AI-TOD, VisDrone, and DTOD), using AdamW optimization with a learning rate of 0.0001 and a batch size of 1.

The multi-expert architecture requires a carefully structured training strategy to ensure stable optimization and effective expert collaboration. We employ a progressive training paradigm where each expert backbone is first pre-trained independently on a combined COCO and AI-TOD dataset to establish stable feature representations. Following individual expert training, multiple experts are progressively integrated through structured grouping, enabling collaborative optimization and parameter alignment across the ensemble. In the final training stage, we construct the complete detection system by selecting the best-performing expert combinations based on validation performance. This progressive approach substantially reduces the training instability common to large ensemble architectures while promoting complementary feature learning among experts. To balance model performance and computational efficiency, only the most effective expert combinations are retained in the final deployment model.

\begin{table*}[!t]
\centering
\begin{tabular*}{\linewidth}{@{\extracolsep{\fill}} l c c c c c c c c @{}}
\toprule
\textbf{Method}   & \textbf{Venue} & \textbf{Backbone}   & \textbf{Param} & $\mathbf{AP_{50}}$ & $\mathbf{AP_{75}}$ & $\mathbf{AP_{50}^S}$ & $\mathbf{AP_{50}^M}$ & $\mathbf{AP_{50}^L}$ \\
\midrule
Faster R-CNN \cite{ren2017fasterrcnn}     & NeurIPS'15     & ResNet-50           & 41M               & 11.6               & 2.1                & 14.8                 & 9.0                  & 22.5                 \\
RetinaNet \cite{lin2020retina}        & ICCV'17        & ResNet-50           & 31M               & 5.2                & 0.7                & 8.9                  & 4.8                  & 8.7                  \\
Cascade R-CNN \cite{cai2018cascade}    & CVPR'18        & ResNet-50           & 68M               & 9.9                & 1.7                & 12.9                 & 8.0                  & 20.7                 \\
YOLOv3 \cite{yolov3}           & ArXiv'18       & DarkNet-53          & 61M               & 4.4                & 0.1                & 5.6                  & 4.2                  & 14.1                 \\
FCOS \cite{fcos}             & ICCV'19        & ResNet-50           & 31M               & 7.0                & 1.2                & 9.3                  & 5.6                  & 10.7                 \\
RepPoints \cite{reppoints}        & ICCV'19        & ResNet-50           & 31M               & 5.0                & 0.8                & 7.2                  & 5.1                  & 10.8                 \\
TridentNet \cite{tridentnet}       & ICCV'19        & ResNet-50           & 32M               & 2.3                & 1.0                & 3.3                  & 3.4                  & 10.2                 \\
ATSS \cite{zhang2020atss}             & CVPR'20        & ResNet-50           & 31M               & 14.6               & 2.2                & 14.0                 & 11.9                 & 29.7                 \\
YOLOv5 \cite{yolov5}           & CVPR'20        & CSPDarkNet-s        & 26M               & 13.8               & 2.4                & 17.6                 & 10.3                 & 29.2                 \\
FoveaBox \cite{foveabox}         & TIP'20         & ResNet-50           & 36M               & 12.0               & 1.8                & 12.7                 & 10.2                 & 28.0                 \\
TPH-YOLOv5 \cite{zhu2021tphyolov5}       & ICCV'21        & CSPDarkNet-s        & 27M               & 10.1               & 1.9                & 14.8                 & 7.0                  & 24.8                 \\
Dyhead \cite{dyhead}           & CVPR'21        & ResNet-50           & 38M              & 10.5               & 1.6                & 12.1                 & 8.2                  & 24.5                 \\
YOLOv8 \cite{yolov8}           & TGRS'22        & CSPDarkNet-s        & 12M               & 10.0               & 1.7                & 12.2                 & 7.5                  & 23.1                 \\
RT-DETR \cite{rtdetr}          & 2023           & HGNetV2             & 31M               & 7.6                & 0.6                & 9.4                  & 5.4                  & 18.7                 \\
DDOD \cite{ddod}             & TMM'23         & ResNet-50           & 32M              & 14.5               & 2.2                & 14.9                 & 11.9                 & 28.6                 \\
DINO \cite{zhang2023dino}             & ICLR'23        & ResNet-50           & 47M              & 8.5                & 1.1                & 12.2                 & 7.0                  & 12.6                 \\
KLDNet \cite{kldnet}           & TGRS'24        & ResNet-50           & 45M               & 13.3               & 2.1                & 13.4                 & 11.3                 & 27.9                 \\
YOLOv10 \cite{yolov10}          & 2024           & CSPDarkNet-s        & 7M                & 11.9               & 2.3                & 13.6                 & 9.2                  & 20.5                 \\
YOLOv11 \cite{yolov11}          & 2024           & C3k2Net-s           & 8M                & 12.2               & 2.2                & 13.9                 & 9.5                  & 23.8                 \\
SCDNet \cite{scdnet}           & TGRS'23        & CSPDarkNet-s        & 43M               & 24.2               & 3.3                & 21.0                 & 18.3                 & 43.7                 \\
\midrule
\textbf{ScaleBridge-Det (Ours)} & -        & Multi-Expert        & 2B              & \textbf{29.8}      & \textbf{7.2}       & \textbf{24.0}        & \textbf{26.0}        & \textbf{42.0}        \\
\bottomrule
\end{tabular*}
\caption{Comparison with state-of-the-art object detection methods on the DTOD dataset. $AP_{50}$ denotes the average precision of all categories at IoU = 0.50. $AP_{75}$ denotes the average precision of all categories at IoU = 0.75. $AP_{50}^S$, $AP_{50}^M$, and $AP_{50}^L$ denote the average precision of small, medium, and large objects at IoU = 0.50, respectively.}
\label{tab:dtod_sota_comparison}
\end{table*}

\begin{table*}[!t]
\centering
\begin{tabular*}{\linewidth}{@{\extracolsep{\fill}} l c c c c c c c c c @{}}
\toprule
\textbf{Method} & \textbf{Year} & \textbf{Backbone} & \textbf{Param} & \textbf{AP} & \textbf{AP50} & \textbf{APvt} & \textbf{APt} & \textbf{APs} & \textbf{APm} \\
\midrule
\multicolumn{10}{l}{\textit{General Object Detection Methods}} \\
RetinaNet \cite{lin2020retina}      & 2020 & ResNet-50 & 33M    & 8.9  & 24.2 & 2.7  & 8.4  & 13.1 & 20.2 \\
Faster-RCNN \cite{ren2017fasterrcnn}    & 2017 & ResNet-50 & 41M    & 12.8 & 29.9 & 0.0  & 9.2  & 24.6 & 37.0 \\
Cascade R-CNN \cite{cai2018cascade}   & 2018 & ResNet-50 & 70M    & 15.1 & 34.2 & 0.1  & 11.5 & 26.7 & 38.5 \\
DetectoRS \cite{qiao2021detectors}    & 2021 & ResNet-50 & 208M   & 16.1 & 35.5 & 0.1  & 12.6 & 28.3 & 40.0 \\
ATSS \cite{zhang2020atss}             & 2020 & ResNet-50 & 32M    & 18.3 & 43.7 & 5.2  & 17.8 & 25.4 & 35.1 \\
GFL \cite{li2020gfl}                  & 2020 & ResNet-50 & 32M    & 19.5 & 47.2 & 6.8  & 19.1 & 26.7 & 36.8 \\
DETR \cite{carion2020detr}                        & 2020 & ResNet-50 & 41M    & 18.2 & 45.3 & 7.5  & 17.6 & 24.8 & 33.2 \\
Deformable DETR \cite{zhu2021deformabledetr}             & 2021 & ResNet-50 & 40M    & 22.6 & 55.1 & 10.3 & 22.1 & 28.5 & 37.4 \\
DAB-DETR \cite{liu2022dabdetr}                    & 2022 & ResNet-50 & 44M    & 23.8 & 57.6 & 11.2 & 23.4 & 29.8 & 38.9 \\
DN-DETR \cite{li2022dndetr}                     & 2022 & ResNet-50 & 44M    & 24.5 & 59.2 & 11.8 & 24.1 & 30.5 & 39.6 \\
DINO \cite{zhang2023dino}                        & 2023 & ResNet-50 & 40M    & 25.9 & 61.3 & 12.7 & 25.3 & 32.0 & 39.7 \\
Group DETR \cite{chen2022groupdetr}                  & 2022 & ResNet-50 & 43M    & 26.4 & 62.8 & 13.1 & 25.9 & 32.6 & 40.3 \\
CoDETR (baseline) \cite{liu2023codetr}           & 2023 & ResNet-50 & 64M    & 28.4 & 66.4 & 12.8 & 29.1 & 30.6 & 45.4 \\
YOLOv8-M \cite{yolov8}                & 2024 & CSPDarkNet & 52M    & 30.1 & 62.1 & 10.6 & 28.4 & 40.2 & 46.5 \\
YOLOv11-M \cite{yolov11}              & 2024 & CSPDarkNet & 48M    & 30.5 & 63.5 & 10.3 & 29.7 & 40.5 & 45.5 \\
YOLOv12-M \cite{yolov12}              & 2024 & CSPDarkNet & 45M    & 26.3 & 56.4 & 7.9  & 24.6 & 35.2 & 40.5 \\
\midrule
\multicolumn{10}{l}{\textit{Remote Sensing Object Detection Methods}} \\
DotD \cite{xu2022dotd}                & 2021 & ResNet-50 & 41M    & 20.4 & 51.4 & 8.5  & 21.1 & 24.6 & 30.4 \\
NWD-RKA \cite{wang2023nwd}            & 2022 & ResNet-50 & 68M    & 22.2  & 52.5 & 7.8  & 21.8 & 28.0 & 37.2 \\
RFLA \cite{zhou2023rfla}              & 2022 & ResNet-50 & 26M    & 25.7 & 58.9 & 9.2  & 25.5 & 30.2 & 40.2 \\
QueryDet \cite{chen2022querydet}                    & 2022 & ResNet-50 & 45M    & 26.8 & 60.4 & 12.5 & 26.3 & 33.1 & 41.5 \\
TPH-YOLOv5 \cite{zhu2021tphyolov5}                  & 2021 & CSPDarkNet & 28M    & 25.3 & 58.7 & 11.8 & 24.9 & 31.2 & 39.8 \\
DNTR \cite{liu2024dntr}                        & 2024 & ResNet-50 & 128M    & 29.0 & 58.3 & 17.3 & 29.2 & 33.5 & 40.5 \\
CFENet \cite{li2024exploring}                      & 2024 & PoolFormer & 35M    & 30.2 & 63.7 & 12.8 & 30.3 & 36.7 & 41.9 \\
DQ-DETR \cite{huang2024dqdetr}                     & 2024 & ResNet-50 & 63M    & 30.5 & 69.2 & 15.2 & 30.9 & 36.8 & 45.5 \\
D3Q \cite{guo2024d3q}                         & 2024 & ResNet-50 & 49M    & 32.1 & 70.8 & 16.5 & 32.5 & 37.2 & 46.1 \\
LAE-DINO  \cite{pan2025locate}                     & 2025 & Swin-T & 180M    & 27.2 & 63.8 & 11.1 & 26.8 & 33.9 & 45.3 \\
DINOv3-7b-sat \cite{dinov3}                        & 2025 & ViT-7B & 7B     & 27.4 & 56.8 & 9.2  & 26.5 & 34.1 & 45.2 \\
\midrule
\textbf{ScaleBridge-Det (ours)}       & - & Multi-Expert & 2B   & 33.6 & 71.8 & 14.2 & 30.4 & 35.1 & 46.8 \\
\textbf{ScaleBridge-Det (ours)}       & - & Multi-Expert & 3B   & \textbf{35.7} & \textbf{72.1} & \textbf{16.2} & \textbf{34.5} & \textbf{39.5} & \textbf{51.4} \\
\bottomrule
\end{tabular*}
\caption{Performance comparison on AI-TOD-v2 test set. ScaleBridge-Det achieves state-of-the-art performance with balanced detection across all object scales. Param. denotes model parameters (M: millions, B: billions). Best results are in \textbf{bold}.}
\label{tab:performance_comparison}
\end{table*}

\subsection{Comparisons with State-of-the-Art Methods}

We compare ScaleBridge-Det with state-of-the-art methods across three challenging benchmarks to demonstrate its effectiveness and generalization capability. The comprehensive comparison includes both general object detection methods and specialized remote sensing approaches.

\textbf{Performance on DTOD Dataset:} To validate the effectiveness on extremely dense tiny object scenarios, we evaluate on the DTOD dataset. Table~\ref{tab:dtod_sota_comparison} presents the comparison with state-of-the-art methods. ScaleBridge-Det achieves 29.8\% AP50, surpassing the previous best SCDNet (24.2\% AP50) by 5.6 percentage points. The superior performance demonstrates our method's capability in handling dense tiny objects through density-guided dynamic query allocation. Notably, our method maintains balanced performance across different object scales (AP50$^{S}$: 24.0\%, AP50$^{M}$: 26.0\%, AP50$^{L}$: 42.0\%), confirming the generalization of the proposed approach across extreme density scenarios.

\textbf{Performance on AI-TODv2:} Table~\ref{tab:performance_comparison} presents detailed comparison results on the AI-TODv2 dataset. ScaleBridge-Det with 2.0 billion parameters achieves 33.6\% AP, surpassing all baseline methods \cite{feng2023deeplearning}. The 3.0B parameter variant further improves to 35.7\% AP, establishing state-of-the-art performance on AI-TOD. Additionally, our method reaches 71.8\% AP50 with the 2.0B model and 72.1\% with the 3.0B variant, indicating robust performance across all object scales. The primary advancement emerges in handling tiny targets: the 3.0B model achieves 16.2\% APvt and 34.5\% APt, substantially exceeding the CoDETR baseline's 12.8\% APvt and 29.1\% APt. This demonstrates the effectiveness of scale-adaptive expert routing and density-guided query allocation for balanced multi-scale detection.

Notably, despite leveraging 7 billion parameters and pre-trained representations from 493M satellite images, DINOv3-7b-sat achieves only 27.4\% AP, underperforming several lightweight baselines. This limitation stems from computational constraints that restrict trainable parameters to only the final two transformer blocks, preventing effective feature adaptation for tiny object detection. These findings indicate that large foundation models require sufficient trainable capacity and task-specific architectural design to achieve superior performance in specialized domains.

\textbf{Cross-Dataset Evaluation on VisDrone with Category Mapping:} Table~\ref{tab:cross_dataset_results} presents cross-dataset evaluation when training on AI-TOD and testing on VisDrone without fine-tuning. All models use top-575 predictions for fair comparison. DQ-DETR completely fails with 0 predictions. CoDETR shows poor generalization (ResNet50: 11.8\% mAP, Swin-L: 17.6\% mAP). Notably, LAE-DINO achieves 29.5\% mAP and 57.5\% mAP50, demonstrating comparable robustness to ScaleBridge-Det despite moderate single-dataset performance (27.2\% AP in Table~\ref{tab:performance_comparison}). This robustness benefits from large-scale pre-training on diverse remote sensing data (LAE-1M dataset with 1.2M images). ScaleBridge-Det achieves the best performance with 30.7\% mAP and 58.8\% mAP50, demonstrating effective generalization across different datasets. Visual comparison in Figure~\ref{fig:cross_domain_comprehensive} further illustrates these differences with predominantly high-quality detections.

\begin{table}[!htb]
\footnotesize
\centering
\renewcommand\arraystretch{1.5}
\resizebox{\linewidth}{!}{
\begin{tabular}{lccc}
\toprule
\textbf{Method} & \textbf{Backbone} & \textbf{mAP} & \textbf{mAP50} \\
\midrule
DQ-DETR & ResNet50 & 2.7 & 8.5 \\
CoDETR & ResNet50 & 11.8 & 24.6 \\
CoDETR & Swin-L & 17.6 & 40.0 \\
LAE-DINO & Swin-T+BERT & 29.5 & 57.5 \\
\textbf{ScaleBridge-Det (ours)} & \textbf{Multi-Expert} & \textbf{30.7} & \textbf{58.8} \\
\bottomrule
\end{tabular}
}
\caption{Cross-dataset evaluation (trained on AI-TOD, tested on VisDrone without fine-tuning). All models use top-575 predictions for fair comparison. Category mapping: AI-TOD \textit{vehicle} $\rightarrow$ VisDrone \{\textit{car, van, truck, tricycle, bus, motor}\}; AI-TOD \textit{person} $\rightarrow$ VisDrone \{\textit{pedestrian, people}\}. DQ-DETR fails completely. ScaleBridge-Det achieves the best performance.}
\label{tab:cross_dataset_results}
\end{table}

\begin{table}[!htb]
\footnotesize
\centering
\renewcommand\arraystretch{1.5}
\resizebox{1.0\linewidth}{!}{
\begin{tabular}{cc|ccccc}
\toprule
\textbf{REM} & \textbf{DGQ} & \textbf{AP} & \textbf{AP$_\text{vt}$} & \textbf{AP$_\text{t}$} & \textbf{AP$_\text{s}$} & \textbf{AP$_\text{m}$} \\
\midrule
 &   & 31.7 & 13.9 & 31.3 & 38.3 & 50.2\\
\checkmark &   & 34.2 & 14.3 & 32.7 & 43.6 & 57.9 \\
 & \checkmark & 32.6 & 15.7 & 33.1 & 37.3 & 49.8\\
\checkmark & \checkmark & \textbf{36.7} & \textbf{15.8} & \textbf{35.2} & \textbf{46.3} & \textbf{58.3}\\
\bottomrule
\end{tabular}
}
\caption{Ablation Study: Component Contributions on AI-TOD-v2 validation set using 2.0B base model. Results are higher than test set performance due to validation-test split difficulty difference.}
\label{tab:ablation_components}
\end{table}

\begin{figure}[!t]
  \centering
  \includegraphics[width=0.48\textwidth]{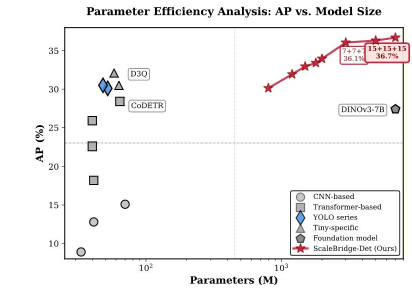}
  \caption{Parameter Efficiency Analysis: AP vs. Model Size on AI-TOD test set. The x-axis (log scale) shows model parameters in millions (M), and the y-axis shows Average Precision (AP\%). Baseline methods are grouped by architecture type: CNN-based (circles), Transformer-based (squares), YOLO series (diamonds), Tiny-specific methods (triangles), and Foundation model (pentagon). The red star-connected line represents ScaleBridge-Det variants with different expert configurations, all using tiny-object-specific pre-training (DIOR + DOTA).}
  \label{tab:expert_ablation_extended}
\end{figure}

\begin{figure*}[!t]
\centering
\includegraphics[height=0.75\textheight,width=\textwidth]{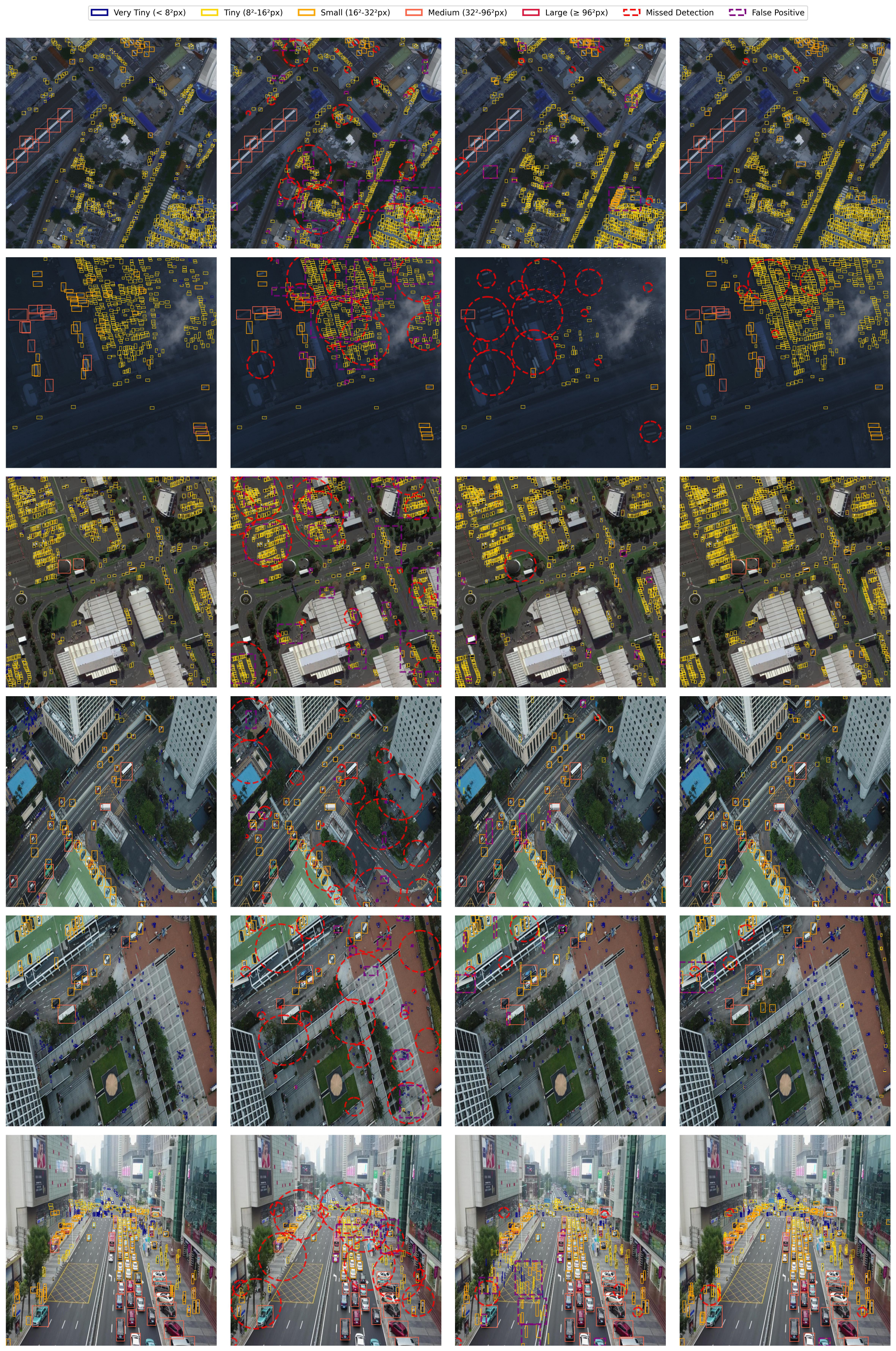}
\caption{Multi-scale detection comparison on challenging scenes. Rows 1-3: AI-TOD; Rows 4-6: VisDrone. Columns: (a) Ground truth, (b) CoDETR, (c) DQ-DETR, (d) ScaleBridge-Det (Ours). Detection boxes are color-coded by scale following Fig.~\ref{fig:scale_distribution}. Red dashed circles mark missed detections; purple dashed rectangles indicate false positives.}
\label{fig:vis_multiscale_detection}
\end{figure*}

\begin{figure*}[!htbp]
\centering
% Top row: DETR, Co-DETR, DQ-DETR
\begin{subfigure}[b]{0.30\textwidth}
    \centering
    \includegraphics[width=\textwidth]{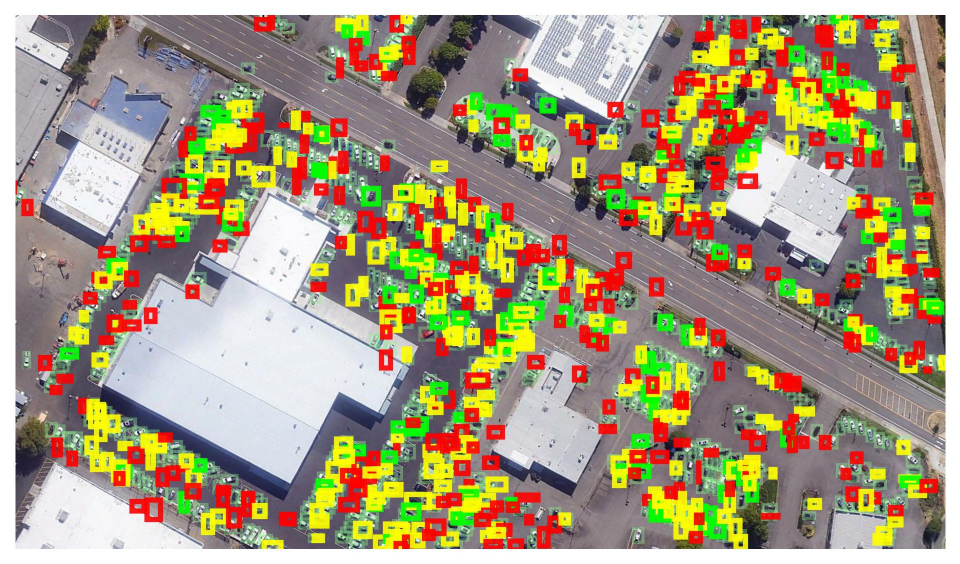}
    \caption{DETR}
    \label{fig:det_detr}
\end{subfigure}
\hfill
\begin{subfigure}[b]{0.30\textwidth}
    \centering
    \includegraphics[width=\textwidth]{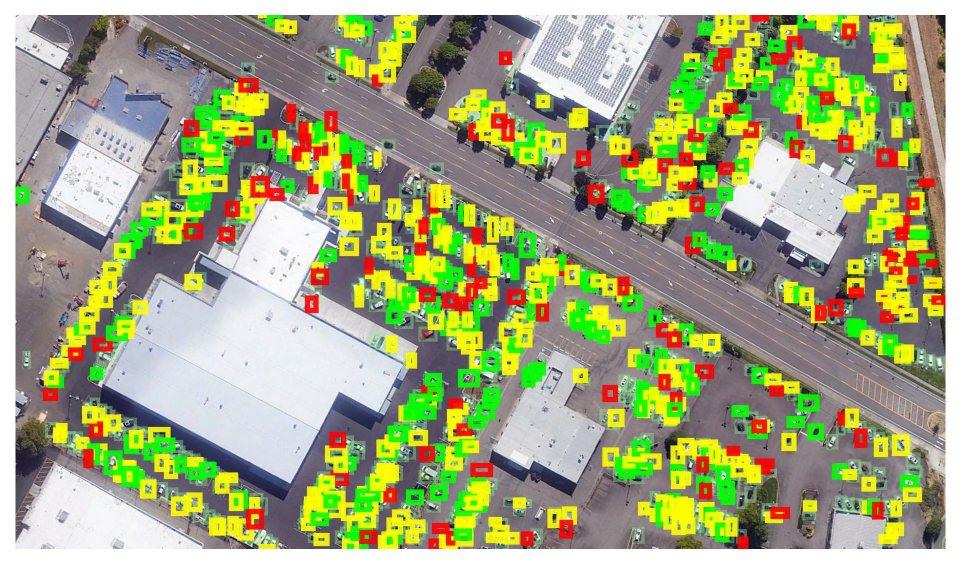}
    \caption{Co-DETR}
    \label{fig:det_codetr}
\end{subfigure}
\hfill
\begin{subfigure}[b]{0.30\textwidth}
    \centering
    \includegraphics[width=\textwidth]{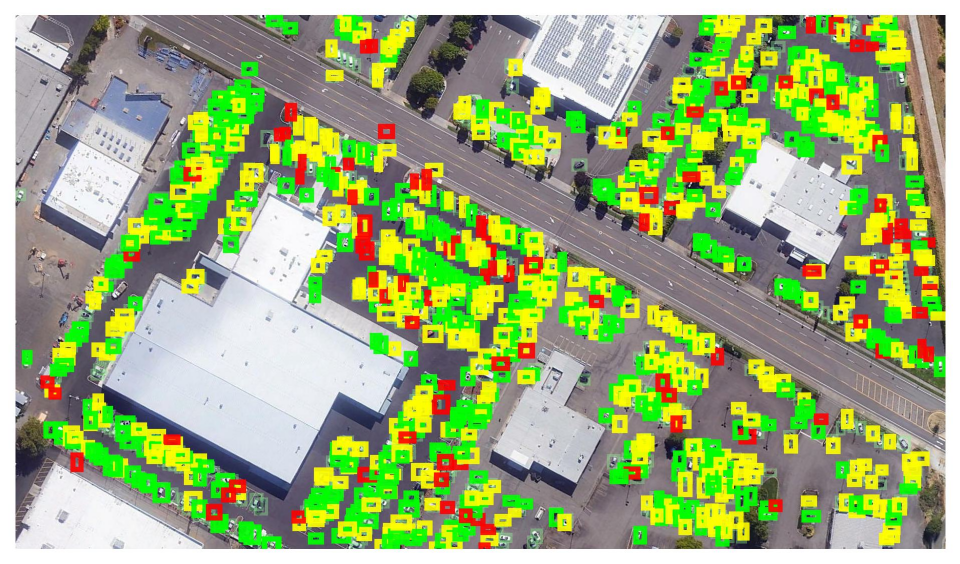}
    \caption{DQ-DETR}
    \label{fig:det_dqdetr}
\end{subfigure}

\vspace{0.10cm}

% Middle row: DINOv3, Ground Truth (larger, centered), DINO-X
\begin{subfigure}[b]{0.30\textwidth}
    \centering
    \includegraphics[width=\textwidth]{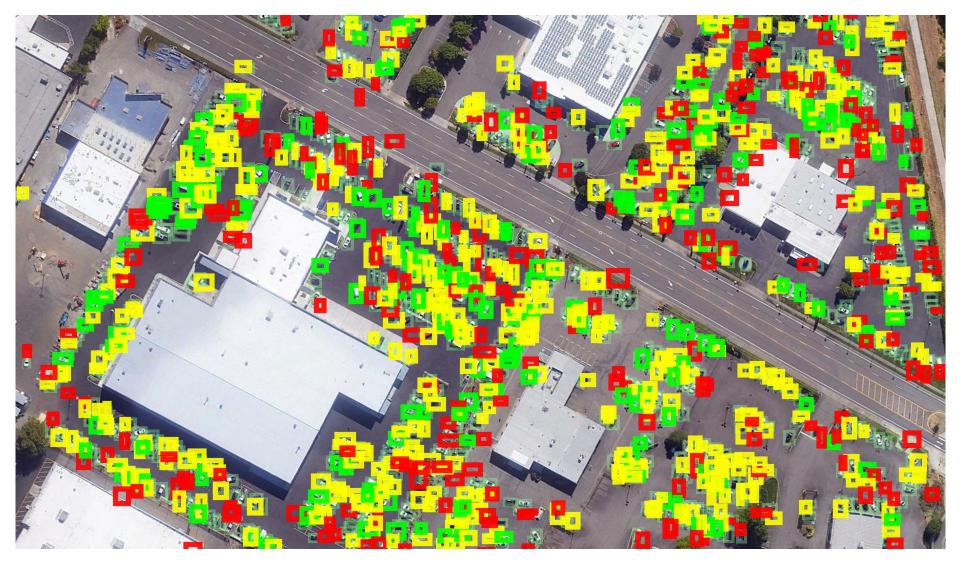}
    \caption{DINOv3}
    \label{fig:det_dinov3}
\end{subfigure}
\hfill
\begin{subfigure}[b]{0.36\textwidth}
    \centering
    \fbox{\includegraphics[width=0.98\linewidth]{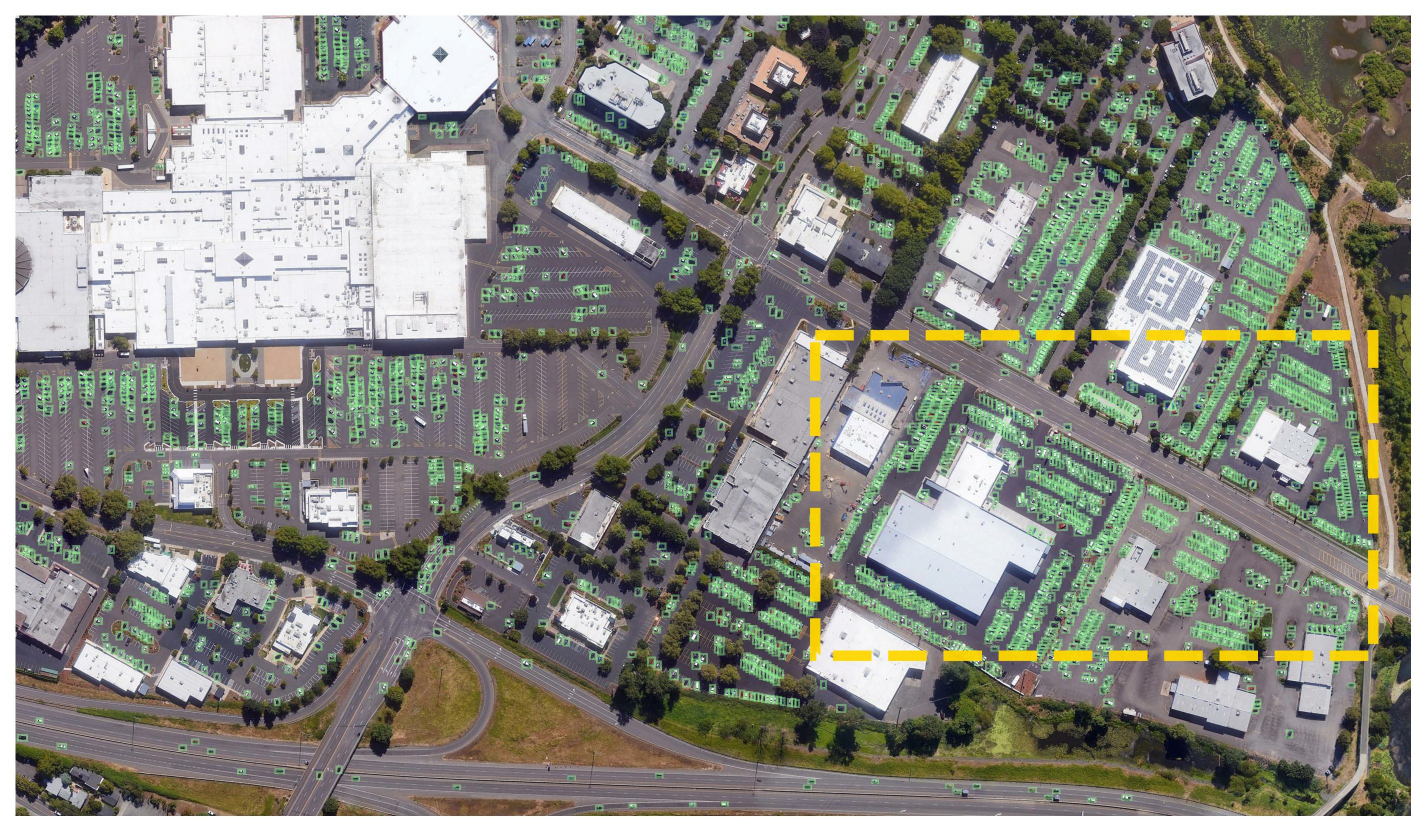}}
    \caption{\textbf{Ground Truth (Full Image)}}
    \label{fig:det_gt}
\end{subfigure}
\hfill
\begin{subfigure}[b]{0.30\textwidth}
    \centering
    \includegraphics[width=\textwidth]{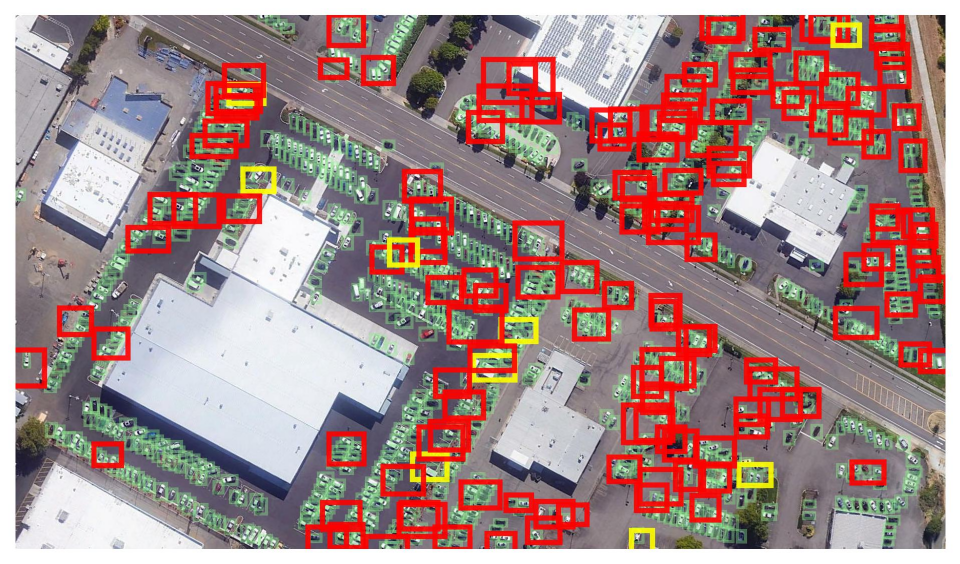}
    \caption{DINO-X}
    \label{fig:det_dinox}
\end{subfigure}

\vspace{0.10cm}

% Bottom row: YOLOv8, YOLOv11, Our Model
\begin{subfigure}[b]{0.30\textwidth}
    \centering
    \includegraphics[width=\textwidth]{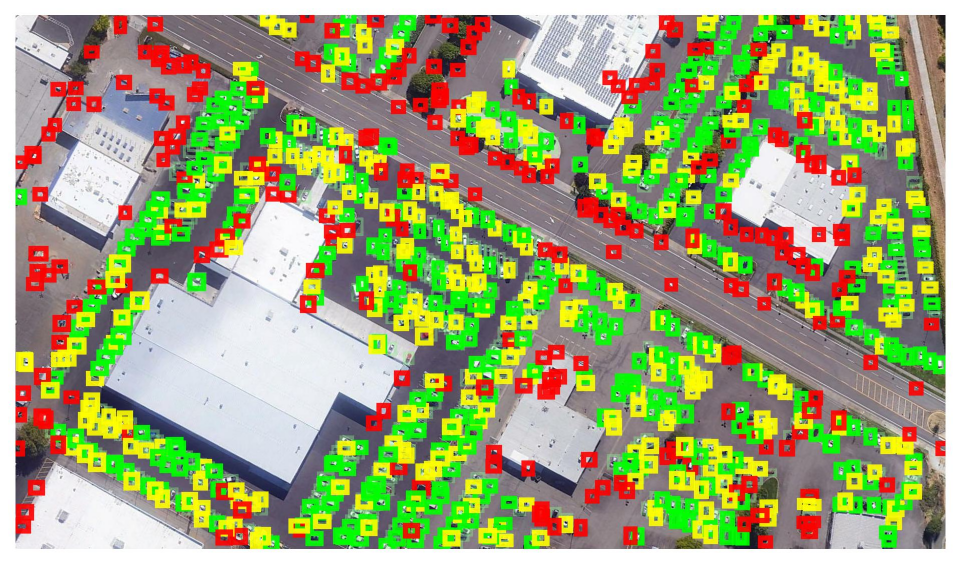}
    \caption{YOLOv8}
    \label{fig:det_yolov8}
\end{subfigure}
\hfill
\begin{subfigure}[b]{0.30\textwidth}
    \centering
    \includegraphics[width=\textwidth]{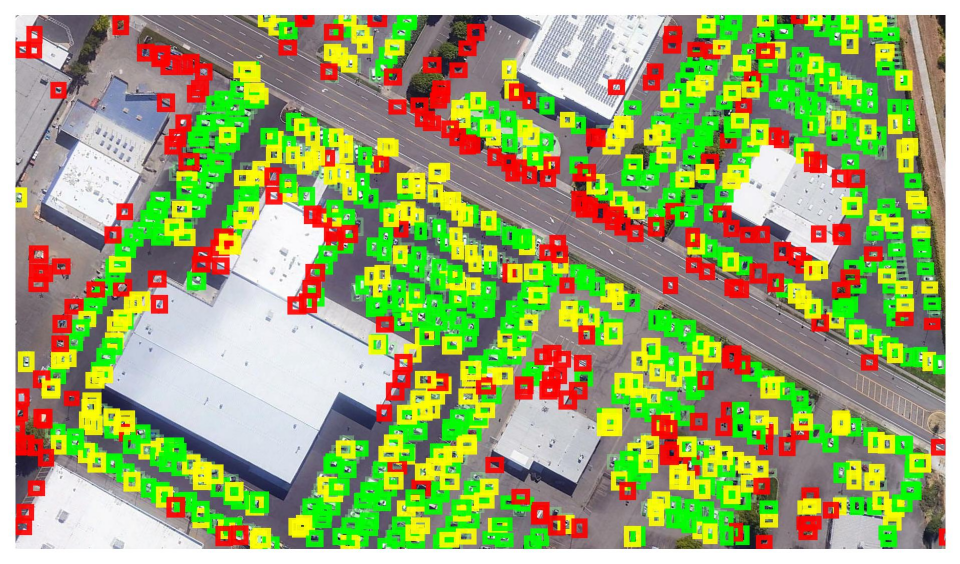}
    \caption{YOLOv11}
    \label{fig:det_yolov11}
\end{subfigure}
\hfill
\begin{subfigure}[b]{0.30\textwidth}
    \centering
    \fbox{\includegraphics[width=0.98\linewidth]{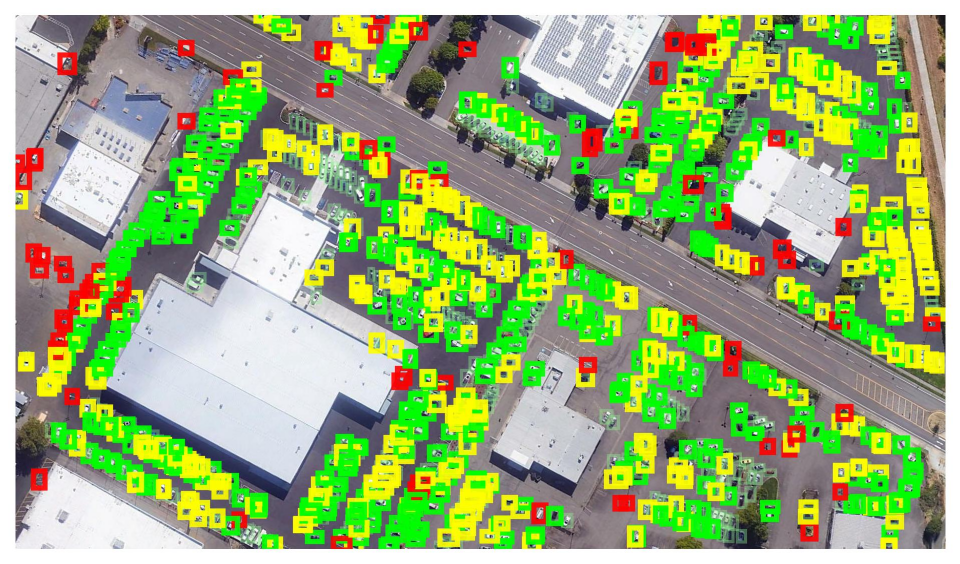}}
    \caption{\textbf{ScaleBridge-Det (Ours)}}
    \label{fig:det_ours}
\end{subfigure}

\caption{Detection comparison on an extremely dense scene (3058 objects). Center (e): \textbf{Ground Truth} with golden dashed box marking the zoom region. Surrounding subfigures show zoomed detection results from different methods. Light green boxes: ground truth; colored predictions: \textcolor{green}{green} (IoU$\geq$0.5), \textcolor{yellow}{yellow} (0.2$\leq$IoU$<$0.5), \textcolor{red}{red} (IoU$<$0.2).}
\label{fig:dense_detection_comparison}
\end{figure*}

\subsection{Ablation Study}

To investigate the contribution of each module in our proposed architecture, we conduct comprehensive ablation experiments on the AI-TOD validation set. We first analyze parameter efficiency by comparing ScaleBridge-Det variants with different expert configurations against state-of-the-art baselines in Figure~\ref{tab:expert_ablation_extended}. The visualization employs different marker shapes to distinguish baseline method categories: CNN-based detectors (circles), Transformer-based approaches (squares), YOLO series (diamonds), Tiny-specific methods (triangles), and Foundation models (pentagons), providing clear visual differentiation across diverse architectural paradigms. The results demonstrate that our multi-expert architecture achieves superior parameter efficiency: with 800M parameters (1+1+1 configuration using ResNet+ViT+Swin experts), ScaleBridge-Det already surpasses most baseline methods. The performance scales consistently with model capacity, exhibiting a smooth growth trajectory from 30.2\% AP at 800M parameters to 36.8\% AP at 7B parameters, where each expert combination contributes complementary features for balanced multi-scale detection. All ScaleBridge-Det variants utilize tiny-object-specific pre-training on DIOR and DOTA datasets, which proves essential for capturing fine-grained features characteristic of remote sensing imagery. Notably, at the same 7B parameter scale, our method substantially outperforms the foundation model DINOv3-7B-sat (36.8\% vs 27.4\% AP, a 9.4 percentage point improvement), demonstrating that task-specific multi-expert architectural design with progressive training surpasses generic foundation model pre-training for specialized tiny object detection tasks.

Table~\ref{tab:ablation_components} shows the performance under different combinations of the Routing-Enhanced Mixture Attention (REM) module and Density-Guided Dynamic Query (DGQ) module.

\textbf{Component Contributions:} The REM module independently contributes a notable 2.5 percentage point AP improvement over the baseline (from 31.7\% to 34.2\%), demonstrating the substantial effectiveness of scale-adaptive expert routing and multi-scale feature fusion. The DGQ module shows particular strength in improving very tiny object detection, with a 1.8 percentage point APvt improvement (from 13.9\% to 15.7\%). The full system combining both modules achieves 36.7\% AP with 15.8\% APvt, representing a 5.0 percentage point overall gain and 1.9 percentage point improvement on very tiny targets.

\textbf{Synergistic Effects:} Notably, the combined performance exceeds the sum of individual contributions, indicating strong synergistic effects between scale-adaptive fusion and density-guided allocation. This interaction particularly benefits tiny objects, reflecting the complementary nature of our proposed mechanisms for balanced multi-scale detection.

\textbf{Implementation Details:} For REM ablation, we replace the integrated routing and hybrid attention with a single Swin-L backbone and standard FPN. For DGQ ablation, we employ fixed query mechanisms with 900 object queries, eliminating density map generation and dynamic allocation. This systematic approach ensures accurate attribution of performance improvements to specific architectural components.

% 1. 检测结果对比（最重要）- 展示3-4个典型场景
% 2. 密度预测热力图（体现DGQ创新）- 2-3个例子
% 3. 注意力/专家激活热力图（体现REM创新）- 1-2个例子

\subsection{Visualization Analysis}

To provide an intuitive understanding of ScaleBridge-Det's effectiveness, a comprehensive qualitative analysis is presented through multi-scale detection visualization, density prediction heatmaps, and expert activation patterns, demonstrating how the framework achieves balanced detection across extreme scale variations while adapting to diverse object density distributions.

Figure~\ref{fig:vis_multiscale_detection} presents a qualitative comparison of detection results on representative scenes from the AI-TOD (rows 1-3) and VisDrone (rows 4-6) datasets, containing extreme scale variations and adverse environmental conditions. The visualization explicitly marks detection failures through red dashed circles (missed detections) and purple dashed rectangles (false positives), revealing critical differences in scale-adaptive capability and robustness across methods. The baseline CoDETR exhibits characteristic weaknesses on small-and-below scale objects, with numerous red circles in column (b) highlighting missed detections of tiny vehicles and small objects, while purple rectangles indicate false positives on small-scale predictions. This demonstrates the tiny-object suppression problem in standard multi-scale detectors that optimize primarily for medium and large objects. While specialized for tiny objects, DQ-DETR shows substantial performance degradation on small-and-above scales, as evidenced by dense red circles marking missed medium/large objects in column (c). Critically, row 2 reveals severe robustness issues under low-light foggy conditions, where DQ-DETR produces extensive missed detections and false positives across the scene, while both CoDETR and ScaleBridge-Det maintain reasonable performance. This indicates that extreme specialization for tiny objects compromises generalization capability and environmental robustness. In contrast, ScaleBridge-Det (column d) demonstrates balanced performance across all scales with minimal marked failures, confirming effective handling of the full scale spectrum from very tiny to large objects while maintaining robustness under challenging illumination and weather conditions.

Figure~\ref{fig:dense_detection_comparison} provides quantitative analysis on a representative DTOD dataset example containing extreme density conditions (3058 objects in a single 4766$\times$2735 scene). DTOD (Dense Tiny Object Detection) specifically targets scenarios with densely packed tiny objects, making it an ideal benchmark for stress-testing detector robustness. The golden dashed box in the center ground truth image highlights a zoom region containing 890 densely packed tiny objects, serving as a critical test case for evaluating detector performance under extreme density. Notably, DINO-X (subfigure f), despite achieving state-of-the-art performance on general object detection benchmarks (COCO, Objects365), exhibits catastrophic failure in this remote sensing scenario with only 108 detections (12\% recall) in the zoom region, demonstrating that models optimized for general domains fail to transfer to remote sensing tiny object detection where object scales and densities fundamentally differ from natural images. Similarly, DINOv3 (subfigure d) and other general-purpose detectors (DETR, YOLOv8, YOLOv11) show extensive missed detections (red boxes), confirming the domain gap between general and remote sensing object detection. CoDETR (subfigure b) and DQ-DETR (subfigure c), while designed for remote sensing, still exhibit substantial missed detections and false positives (yellow/red boxes), as these specialized methods partially address density challenges but lack comprehensive mechanisms to handle the extreme scale-density combinations characteristic of real-world aerial imagery. ScaleBridge-Det (subfigure i) achieves 659 detections (74\% recall) with predominantly green boxes indicating high-quality detections (IoU$\geq$0.5), representing a 6$\times$ improvement over DINO-X and 2-3$\times$ improvement over remote sensing specialists, validating the effectiveness of density-guided query allocation in adapting to extreme object densities while maintaining detection quality through scale-adaptive expert routing.

\section{Conclusion}
This work presents ScaleBridge-Det, a novel framework that addresses the fundamental challenge of balanced detection across extreme scale variations in remote sensing imagery through a simple yet effective approach. The proposed method incorporates a Routing-Enhanced Mixture Attention module that leverages a mixture-of-experts architecture with adaptive routing mechanisms to dynamically integrate scale-specific expert features, effectively resolving the scale competition problem where tiny objects are suppressed by dominant large structures. Additionally, the Density-Guided Dynamic Query module provides spatial awareness of object distribution patterns and adaptively adjusts query allocation according to predicted density maps, enabling efficient resource distribution across varying density scenarios while maintaining computational efficiency. Extensive experiments on challenging benchmarks demonstrate that the framework achieves balanced detection performance across all object scales, from very tiny targets occupying fewer than eight pixels to large structures spanning hundreds of pixels, while exhibiting superior cross-domain generalization capability. Future research will focus on developing lightweight variants suitable for resource-constrained UAV platforms and extending the framework to accommodate additional remote sensing modalities, such as synthetic aperture radar and hyperspectral imagery, further advancing the practical deployment of large detection systems in diverse real-world scenarios.

%\FloatBarrier %强制悬浮设为之前的
\appendices

\ifCLASSOPTIONcaptionsoff
  \newpage
\fi

\bibliographystyle{IEEEtran}
\bibliography{ref}
\end{document}